\pdfoutput=1

\documentclass[11pt]{article}

\usepackage[utf8]{inputenc} 
\usepackage[T2A,T1]{fontenc} 
\usepackage[russian,english]{babel} 

\usepackage[final]{acl}
\usepackage{times}
\usepackage{inconsolata} 
\usepackage{microtype}   

\usepackage[table,dvipsnames]{xcolor} 
\usepackage{graphicx}
\usepackage{pgfplots}
\usepgfplotslibrary{groupplots}
\pgfplotsset{compat=1.18}

\usepackage{amsmath}
\usepackage{amssymb}
\usepackage{latexsym}
\usepackage{pifont} 

\usepackage{booktabs}   
\usepackage{multirow}   
\usepackage{multicol}   
\usepackage{longtable}  
\usepackage{adjustbox}  
\usepackage{caption}
\usepackage{float}

\usepackage{url}
\usepackage[many]{tcolorbox}
\tcbuselibrary{listings}

\newcommand{\cmark}{\textcolor{green!60!black}{\ding{51}}} 
\newcommand{\xmark}{\textcolor{red!70!black}{\ding{55}}}   

\definecolor{darkgreen}{rgb}{0.0,0.5,0.0}

\title{\textsc{MathMist}: A Parallel Multilingual Benchmark Dataset for Mathematical Problem Solving and Reasoning}

\author{
\textbf{Mahbub E Sobhani\textsuperscript{1,2}},
\textbf{Md. Faiyaz Abdullah Sayeedi\textsuperscript{2,4}},
 \textbf{Tasnim Mohiuddin\textsuperscript{3}\textsuperscript{$\dagger$}},
\\
 \textbf{Md. Mofijul Islam\textsuperscript{5,6} \textsuperscript{$\dagger$} \thanks{Work does not relate to position at Amazon.}},
 \textbf{Swakkhar Shatabda\textsuperscript{1}\thanks{Equal supervision.} \thanks{Correspondence: \href{mailto:swakkhar.shatabda@bracu.ac.bd}{swakkhar.shatabda@bracu.ac.bd}}}
\\
\\
 \textsuperscript{1}BRAC University,
 \textsuperscript{2}United International University,
 \textsuperscript{3}Qatar Computing Research Institute, 
 \\
 \textsuperscript{4} Center for Computational \& Data Sciences, Independent University, Bangladesh, \\
 \textsuperscript{5}Amazon GenAI,
 \textsuperscript{6}University of Virginia
\\
\small{
   \href{https://huggingface.co/datasets/mahbubhimel/MathMist}{%
   \includegraphics[height=1.5em]{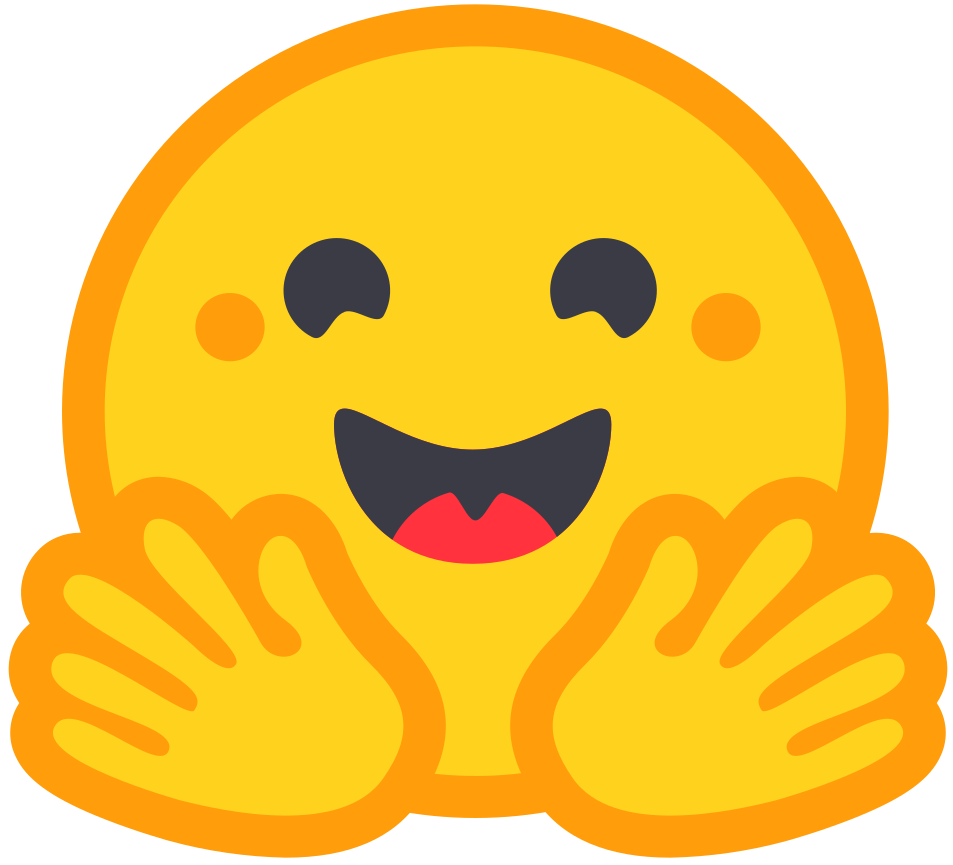}\enspace HuggingFace/Datasets/MathMist}
}
}

\newcommand{\dataname}{\textsc{MathMist}}

\begin{document}
\maketitle
\begin{abstract}
Mathematical reasoning remains one of the most challenging domains for large language models (LLMs), requiring not only linguistic understanding but also structured logical deduction and numerical precision. While recent LLMs demonstrate strong general-purpose reasoning abilities, their mathematical competence across diverse languages remains underexplored. Existing benchmarks primarily focus on English or a narrow subset of high-resource languages, leaving significant gaps in assessing multilingual and cross-lingual mathematical reasoning. To address this, we introduce \textbf{{\dataname}}, a parallel multilingual benchmark for mathematical problem solving and reasoning. {{\dataname}} encompasses 2,890 parallel Bangla-English gold standard artifacts, totaling $\approx$30K aligned question–answer pairs across thirteen languages, representing an extensive coverage of high-, medium-, and low-resource linguistic settings. The dataset captures linguistic variety, multiple types of problem settings, and solution synthesizing capabilities. We systematically evaluate a diverse suite of models, including open-source small and medium LLMs, proprietary systems, and multilingual-reasoning-focused models under zero-shot, chain-of-thought (CoT), perturbated reasoning, and code-switched reasoning paradigms. Our results reveal persistent deficiencies in LLMs’ ability to perform consistent and interpretable mathematical reasoning across languages, with pronounced degradation in low-resource settings. All the codes and data are available at GitHub:  \url{https://github.com/mahbubhimel/MathMist}
\end{abstract}

\section{Introduction}\label{intro}

Mathematical reasoning serves as one of the most rigorous tests of a model’s ability to integrate linguistic understanding with structured logical deduction and quantitative computation. While recent large language models (LLMs) demonstrate impressive capabilities in natural language understanding and general reasoning \citep{deepseekai2025deepseekr1incentivizingreasoningcapability}, their mathematical competence remains uneven, particularly when problems are presented in typologically diverse or low-resource languages \citep{ahn-etal-2024-large}.

\begin{table*}[!t]
\centering
\small
\begin{tabular*}{\textwidth}{l|c|c|c|c|c|c|c}
\hline
\rowcolor{gray!15} \textbf{Dataset} & \textbf{Lang.} & \textbf{Size} & \textbf{MCQ} & \textbf{CoT} & \textbf{CS-CoT} & \textbf{Perturb. Reasoning} & \textbf{Parallel} \\
\hline
BenNumEval \citep{ahmed-etal-2025-bennumeval} & bn & 3.2K & \cmark & \cmark & \xmark & \xmark & \xmark \\
SOMADHAN \citep{paul2025leveraging} & bn & 8.8K & \xmark & \cmark & \xmark & \xmark & \xmark \\
MMATH \citep{luo-etal-2025-mmath} & 10 & 3.7K & \xmark & \cmark & \xmark & \xmark & \cmark \\
MGSM8KInstruct \citep{chen-etal-2024-breaking} & 10 & 8K & \xmark & \cmark & \xmark & \xmark & \cmark \\
ConceptMath \citep{wu-etal-2024-conceptmath} & en, zh & 4.0K & \xmark & \cmark & \xmark & \xmark & \cmark \\
MathQA-TR \citep{gedik2023solving} & en, tr & 37.2K & \xmark & \xmark & \xmark & \xmark & \cmark \\
HAWP \citep{sharma-etal-2022-hawp} & en, hi & 2.3K & \xmark & \cmark & \xmark & \xmark & \cmark \\
ArMATH \citep{alghamdi-etal-2022-armath} & ar & 6K & \cmark & \xmark & \xmark & \xmark & \xmark \\
MATH \citep{hendrycks2021measuring} & en & 12.5K & \xmark & \cmark & \xmark & \xmark & \xmark \\
KoTAB \citep{9070606} & ko & 1.1K & \cmark & \xmark & \xmark & \xmark & \xmark \\
\hline
\rowcolor{blue!10} \textbf{\texttt{\dataname} (Ours)} & \textbf{13} & \textbf{29K+} & \cmark & \cmark & \cmark & \cmark & \cmark \\
\hline
\end{tabular*}
\caption{Comparison of mathematical reasoning datasets. Language codes: bn = Bengali, en = English, zh = Chinese, ar = Arabic, tr = Turkish, ko = Korean, hi = Hindi. 
CS-CoT = Code-Switched Chain-of-Thought, Perturb. = Perturbation.}
\label{tab:data_comparison}
\end{table*}

Existing mathematical reasoning benchmarks, such as MathQA \citep{amini-etal-2019-mathqa} and GSM8K \citep{cobbe2021training}, have catalyzed progress in English-centric evaluation. However, they offer little insight into how LLMs reason in multilingual or cross-lingual contexts. This limitation is critical because mathematical reasoning is not language-agnostic—it depends on the precise linguistic framing of problems, the syntactic and morphological structures of languages, and the semantic mapping of mathematical terms. As a result, evaluating reasoning only in English systematically overlooks how LLMs process and transfer reasoning skills across languages, especially between high-resource (e.g., English, French) and low-resource (e.g., Bangla, Kazakh) linguistic settings.

Mathematical Word Problem (MWP) solving requires not only linguistic understanding but also symbolic reasoning, proof formulation, and generalization across typologically diverse languages \citep{10.1109/TPAMI.2019.2914054}. Yet, existing corpora are either monolingual or fully translation-based, focusing mainly on final-answer accuracy rather than intermediate reasoning quality \citep{ahmed-etal-2025-bennumeval, paul2025leveraging}. They have enriched reasoning evaluation, but are confined to arithmetic tasks. Similarly, multilingual datasets expand coverage but lack parallel structure, synthetic perturbations, and multi-format evaluation \citep{chen-etal-2024-breaking}. Consequently, there remains a significant gap in understanding how LLMs reason mathematically across different linguistic settings and how they respond in code-switched settings or fallacious reasoning.

To address these limitations, we introduce \textbf{\dataname}, a parallel multilingual benchmark designed to evaluate LLMs’ mathematical problem solving and reasoning across a diverse set of languages and task settings. {\dataname} provides a controlled platform to assess how models handle linguistic variation, code-switching, and reasoning perturbations within equivalent mathematical contexts. The dataset encompasses $\approx$30K mathematical artifacts spanning thirteen languages—English, Arabic, Bangla, French, Swahili, Persian, Turkish, Hausa, Gujarati, Amharic, Kazakh, Finnish, and Lithuanian. This selection covers a wide range of high-, medium-, and low-resource languages from various linguistic families, collectively reaching around 3.15 billion speakers (see Appendix \ref{appendix:language_stat}). Our key contributions are as follows:

\begin{itemize}
    \item We introduce {\dataname}, a comprehensive dataset that encompasses
    2,890 parallel Bangla-English gold standard math word problems. It comprises approximately 30K aligned question-answer pairs verified by subject matter experts across thirteen languages. The dataset includes 18.9K parallel problems, 2.2K multiple-choice questions, and 8.6K perturbed solutions, enabling fine-grained evaluation of reasoning behaviors and error sensitivity across various linguistic settings.
    \item We design a suite of task variations enabling diverse reasoning assessment, including code-switched CoT reasoning between high- and low-resource languages and perturbation reasoning through controlled error injection.
    \item We conduct extensive qualitative and quantitative evaluations using state-of-the-art LLMs, analyzing both stepwise reasoning accuracy and cross-lingual generalization.
\end{itemize}

Overall, \dataname \enspace provides the first large-scale, parallel, and linguistically diverse benchmark for mathematical reasoning across multiple languages. By enabling systematic cross-lingual evaluation, it paves the way for a deeper understanding of multilingual reasoning processes in LLMs and establishes a foundation for developing models that reason more reliably, equitably, and transparently across the world’s languages.

\section{Related Work}\label{back}

\begin{figure*}[t!] 
    \centering
    \includegraphics[width=\textwidth]{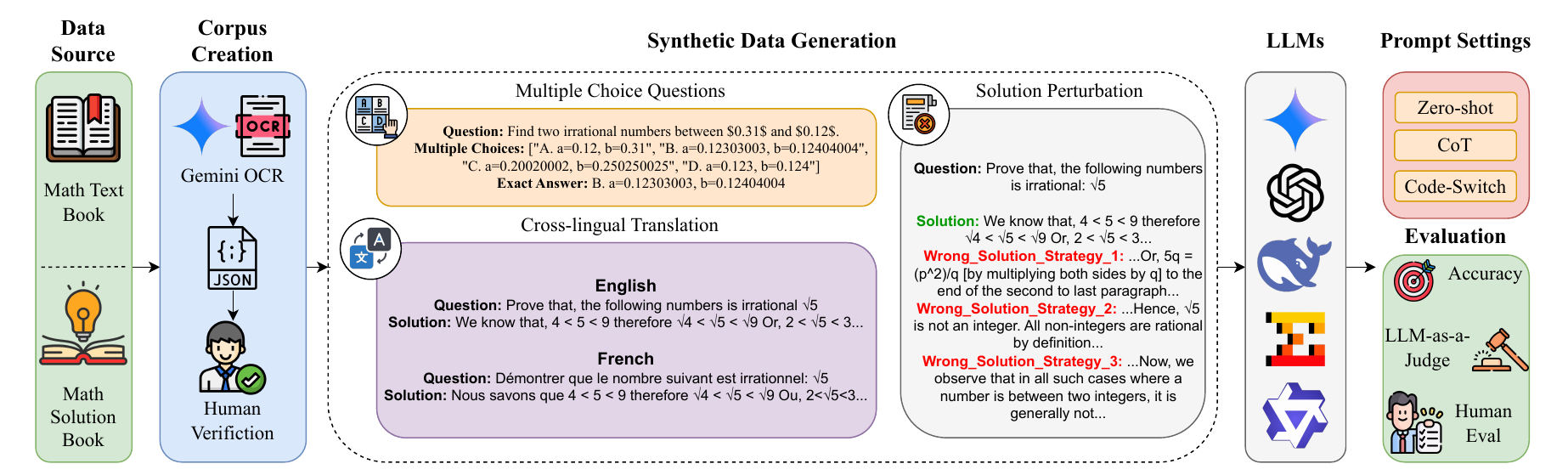}
    \caption{Overview of \textbf{{{\dataname}}} data creation and evaluation pipeline. \textbf{(Left)} Data Sourcing and corpus creation uses \texttt{Gemini} OCR on textbooks, stores data to JSONL, and applies human verification. \textbf{(Center)} Synthetic data generation encompasses \texttt{Multiple Choice Question} (MCQ) generation, \texttt{Cross-Lingual Translation}, and \texttt{Solution Perturbation}. \textbf{(Right)} The evaluation process tests various LLMs under different prompt settings.}
    \label{fig:mathmist_data_pipeline} 
\end{figure*}

\paragraph{Low-Resource Math Datasets.}
Recent research has begun to make measurable progress toward enhancing mathematical reasoning capabilities in low-resource languages. \citet{ahmed-etal-2025-bennumeval} introduced BenNumEval, a benchmark for numerical reasoning in Bengali with six task families and over 3.2k problems, showing that even advanced prompting methods such as Cross-Lingual Prompting (XLP) and Cross-Lingual Chain-of-Thought (XCoT) fall short of human-level performance. \citet{mondal2025bmwp} released BMWP, containing 8,653 Bengali math word problems for operation prediction using deep models, achieving 92\% accuracy. \citet{era2024empowering} created PatiGonit, which contains 10k problems for equation translation and found that transformer models like mT5 \citep{xue2021mt5} reached 97.3\% accuracy. Building on these, \citet{paul2025leveraging} introduced SOMADHAN, containing 8,792 manually annotated math word problems with step-by-step reasoning, showing that few-shot Chain-of-Thought prompting improved accuracy up to 88\% with LLaMA-3.3-70B. Despite these efforts, most Bengali datasets are limited to arithmetic problems and final-answer evaluation. Beyond Bengali, \citet{mahgoub2024mathematical} proposed a Synthetic Data Augmentation framework for Arabic mathematical problem solving, showing improvements under data scarcity.

\paragraph{Multilingual Mathematical Reasoning.}
Multilingual studies have explored how cross-lingual training and data balance affect reasoning. \citet{chen-etal-2024-breaking} introduced MGSM8KInstruct, a dataset spanning 10 languages, demonstrating that multilingual supervised fine-tuning improves both cross-lingual and monolingual reasoning. \citet{zhang2024enhancing} further showed that self-distillation from high-resource to low-resource languages enhances accuracy while reducing data and translation costs. More recent benchmarks have moved beyond grade-school arithmetic: \citet{luo-etal-2025-mmath} proposed MMATH, a multilingual benchmark for complex mathematical reasoning across 10 languages, highlighting language inconsistency and off-target reasoning issues, while \citet{wang2025polymath} introduced PolyMath to evaluate multilingual mathematical reasoning robustness across diverse linguistic contexts. In parallel, \citet{wu-etal-2024-conceptmath} introduced ConceptMath, a bilingual (English--Chinese) concept-wise benchmark enabling fine-grained analysis of reasoning performance across hierarchical mathematical concepts. These findings highlight the importance of balanced multilingual corpora and adaptive fine-tuning. Moreover, LLMs are increasingly used for education and tutoring involving mathematical reasoning. \citet{tonga2025simulating} simulated multilingual LLM tutoring and found that native-language feedback improves reasoning, especially in low-resource contexts. \citet{mahran2025investigating} developed a multilingual pipeline for generating and grading math problems, revealing persistent linguistic bias favoring English outputs.

Across prior work, low-resource languages have received limited attention. Also, existing datasets focus mainly on arithmetic problems and lack symbolic and proof-based reasoning tasks. They also emphasize final answers rather than intermediate reasoning, even in recent multilingual benchmarks \citep{wu-etal-2024-conceptmath,luo-etal-2025-mmath,son-etal-2025-linguistic}. Our work addresses these gaps by introducing a comprehensive multilingual mathematical reasoning benchmark with decision-aware, answer-type–specific evaluation, offering fine-grained assessment of reasoning performance. The comparison of mathematical reasoning datasets across multiple languages is shown in Table \ref{tab:data_comparison}.

\section{Corpus Creation}\label{corpus}



We introduce {{\dataname}}, a benchmark dataset developed for multilingual reasoning, which also features a strategic three-stage extension to further challenge LLMs and present deeper insights. The complete {{\dataname}} creation pipeline is shown in Figure \ref{fig:mathmist_data_pipeline}.


\subsection{Data Sourcing}
Primarily, {{\dataname}} contains a total of 2,890 gold standard Bangla-English parallel math word problems, with 1,445 problems available in both Bangla and English. The problems were sourced from the {National Curriculum \& Textbook Board} ({NCTB}) secondary school mathematics books for the 2018–2019 academic year \footnote{https://nctb.gov.bd/} \footnote{https://www.panjeree.com/}, which had more exemplary content than recent editions. We excluded chapters requiring diagrammatic reasoning from the Mathematics book (chapters 6, 7, 8, and 15) and the Higher Mathematics book (chapters 3, 4, and 11), focusing only on those answerable without illustrations. All included examples and exercises were authored and verified by mathematics experts appointed by the People's Republic of Bangladesh. We intentionally kept variable names and mathematical expressions wrapped up with LaTeX to facilitate future automated processing.

Eight second-year undergraduate students volunteered as data collectors to extract questions and solutions by taking screenshots from source books. They were selected for their recent completion of high-school mathematics, ensuring a reliable understanding of the materials. We used \verb|Gemini 2.0 Pro| for OCR to transcribe both Bangla and English mathematical text and LaTeX expressions (see Appendix \ref{appendix:prompts}). Students verified the OCR output against the actual example, making corrections and flagging any errors for re-inspection. This process yielded 1,445 examples, with only 87 requiring manual correction and re-inspection. In the second iteration, each example was tagged as Numerical, Symbolic, or Proof, and the parallelism between Bangla and English was re-verified. A third-year student conducted an additional alignment check, after which five of the authors conducted a final, thorough review to confirm that each instance and its label were accurately aligned and identical to the source. 

\subsection{Multiple-Choice Questions Generation}
To facilitate multiple-choice generation, we employed a distractor-generation strategy that produces confusing but verifiably incorrect options. For a problem with correct solution $A$, we sample $k$ distractors from three options, $\mathcal{D}=\operatorname{sample}_k\!\big(\mathcal{D}_{\mathrm{calc}}\cup\mathcal{D}_{\mathrm{concept}}\cup\mathcal{D}_{\mathrm{plaus}}\big)$, where $\operatorname{sample}_k$ selects $k$ distractors according to mixture weights $(p_{\mathrm{calc}},p_{\mathrm{concept}},p_{\mathrm{plaus}})$ with $p_{\cdot}\ge 0$ and $\sum p_{\cdot}=1$. Representative constructions include $\mathcal{D}_{\mathrm{calc}}\ni\{A\pm1,-A,10A,A/10\}$ (off-by-one, sign, or decimal errors), $\mathcal{D}_{\mathrm{concept}}\ni\{f_{\mathrm{wrong}}(x)\mid f_{\mathrm{wrong}}\neq f_{\mathrm{true}}\}$ (incorrect formula or unit confusion), and $\mathcal{D}_{\mathrm{plaus}}\ni\{A+\delta,\mathrm{round}(A,r)\}$ (near-miss values from small perturbations or rounding). We enforce $D\neq A$ for all $D\in\mathcal{D}$ and verify candidates with an automatic verifier $V(\cdot)$ such that $V(D)=\mathtt{false}$. Moreover, we require $|D-A|>\tau$ to avoid numerical ties. Each item is stored as the tuple $(Q,A,\mathcal{D}_{en})$ and $(Q,A,\mathcal{D}_{bn})$ to ensure all multiple choices are identical in both languages. These annotations enable later analysis of distractor utility and targeted evaluation of model weaknesses.

\subsection{Perturbation Generation}\label{perturb_sub}
In our perturbation-injection pipeline, each item is denoted as \((Q, S_{\text{true}}, A)\) with one of three strategies from \(\Sigma = \{\sigma_1, \sigma_2, \sigma_3\}\), where \(S_{\text{true}}\) represents the original correct solution. The strategy \(\sigma_1\) infiltrates \{\texttt{step omission}, \texttt{incorrect rule}, \texttt{faulty causality}\}, \(\sigma_2\) insinuates \{\texttt{overgeneralization}, \texttt{logical fallacy}\}, and \(\sigma_3\) includes all five fallacy types. For each item, we forge \(\tilde{S} = \sigma_i(S_{\text{true}})\) while ensuring that the perturbed solution keeps the tone and structure of \(S_{\text{true}}\), with errors seamlessly embedded. Each instance \((Q, A, \tilde{S}, \text{strategy} = \sigma_i)\) must pass automated quality checks by a language model. For the English–Bangla parallel corpus, we ensure version-wise equivalence, given \((S_{\text{en}}, S_{\text{bn}})\) and \(\sigma_i\), we generate \((\tilde{S}_{\text{en}}, \tilde{S}_{\text{bn}})\) while preserving error types and their locations. Furthermore, we verified the quality of error injections with subject matter experts (SMEs) to enable fair cross-lingual evaluation.


\subsection{Multilingual Translation Pipeline}
To evaluate LLM families on MWP solving across diverse resource levels while ensuring typological variety, we selected high-resource languages such as English, French, and Arabic (Indo-European, Semitic); medium-resource languages like Turkish (Turkic), Persian (Indo-Iranian), Swahili (Bantu), Gujarati (Indo-Aryan), Finnish (Uralic), Lithuanian (Indo-European), and low-resource languages, including Bangla (Indo-Aryan), Hausa (Afroasiatic), Amharic (Semitic), and Kazakh (Turkic). With this selection, we augmented our dataset by embodying systematic mathematical translations. Each item in the dataset is denoted as \((Q, S_{\mathrm{true}}, A)\). For the target language set $L$ = \{\texttt{Arabic}, \texttt{French}, \texttt{Swahili}, \texttt{Persian}, \texttt{Turkish}, \texttt{Hausa}, \texttt{Gujarati}, \texttt{Amharic}, \texttt{Kazakh}, \texttt{Finnish}, \texttt{Lithuanian}\}, language-specific translator agents \(\mathcal{A}_\ell\) generate candidate translations \((Q_\ell, S_\ell, A_\ell)\) using a zero-shot prompt. Each candidate is evaluated by a verifier LLM \(V_\ell\) based on criteria \(\mathcal{Q} = \{\mathcal{M}, \mathcal{T}, \mathcal{L}, \mathcal{C}\}\), which comprise mathematical fidelity, terminological correctness, clarity, and completeness. A candidate is accepted if they meet all criteria. If any standards fails, the verifier \(V_\ell\) produces a corrected translation \((\tilde{Q_{\ell}},\tilde{S_{\ell}},\tilde{A_{\ell}})=V_{\ell}(Q_{\ell}, S_{\ell}, A_{\ell}; Q, S_{\mathrm{true}}, A)\), which must satisfy all the criteria. To further assess the translation quality, we engaged subject-matter experts (SMEs) for back-translations using Google Translate \citep{google_translate} and DeepL \citep{deepl_translate}. If the translations did not match the original English, we provided the candidate translations with explicit error points to Gemini for further refinement. We achieved accurate translations from Gemini on the first attempt due to its strong language coverage. The LLM-SME pipeline maintains isomorphism with the original, creating a robust benchmark MWP dataset for mathematical reasoning across diverse resource languages.

\subsection{Corpus Statistics}
The multilingual corpus consists of 1,445 math problems for each of these languages: Bangla, English, French, Kazakh, Finnish, Lithuanian, Turkish, Persian, Arabic, Swahili, Hausa, Gujarati, and Amharic, totaling 18,785 problem instances. It comprises 10,959 Numerical problems (58.34\%), 3,770 Symbolic problems (20.07\%), and 4,056 Proof problems (21.59\%). Within the Numerical category, arithmetic and algebra are most common, with 2,142 and 1,386 instances per language, respectively. Furthermore, most Symbolic problems are algebraic, summing 1,610 math problems per language. Measurements and algebra are dominant in the Proof category. Table \ref{tab:corpus_stat} shows the statistics for one representative language, as all languages share the same distribution. Additionally, 2,266 multiple-choice questions (MCQs) were created for Bangla and English, along with 8,670 incorrect solution variants. In total, the corpus consists of 29,721 artifacts: 18,785 problems, 2,266 MCQs, and 8,670 perturbed solutions. {\dataname} facilitates multilingual modeling of problem comprehension and evaluates LLMs' solving capability.
\begin{table}[t]
\centering
\small
\begin{tabular}{llc}
\toprule
\rowcolor{gray!15} \textbf{Category} & \textbf{Subdomain} & \textbf{Count} \\
\midrule
\multirow{7}{*}{\textbf{Numerical}} 
 & Algebra & 198 \\
 & Arithmetic & 306 \\
 & Measurements & 28 \\
 & Trigonometry & 115 \\
 & Word Problem & 103 \\
 & Probability & 23 \\
 & Series & 70 \\
 & \textbf{Total (Numerical)} & \textbf{843} \\
\midrule
\multirow{3}{*}{\textbf{Symbolic}} 
 & Algebra & 230 \\
 & Measurements & 60 \\
 & \textbf{Total (Symbolic)} & \textbf{290} \\
\midrule
\multirow{6}{*}{\textbf{Proof}} 
 & Algebra & 74 \\
 & Measurements & 160 \\
 & Combinatorics & 32 \\
 & Number Theory & 15 \\
 & Others & 31 \\
 & \textbf{Total (Proof)} & \textbf{312} \\
\midrule
\multicolumn{2}{l}{\textbf{Total per Language}} & \textbf{1,445} \\
\multicolumn{2}{l}{\textbf{Across 13 Languages}} & \textbf{$1,445 \times 13 = 18,785$} \\
\midrule
\multicolumn{2}{l}{\textbf{MCQs (BN \& EN)}} & \textbf{2,266} \\
\multicolumn{2}{l}{\textbf{Perturbed Solutions (BN \& EN)}} & \textbf{8,670} \\
\midrule
\rowcolor{blue!10} \multicolumn{2}{l}{\textbf{Grand Total Artifacts}} & \textbf{29,721} \\
\bottomrule
\end{tabular}
\caption{Overall corpus statistics across all thirteen languages in \texttt{\dataname}.}
\label{tab:corpus_stat}
\end{table}
\section{Experimental Setup}

Our experimental setup includes model configuration, evaluation metrics, and prompt techniques.

\subsection{Models} 

We evaluated state-of-the-art LLMs ranging from \textbf{7B--20B} parameters, covering both open-source and proprietary families. The assessment spans high- to low-resource languages, varied question structures, and code-switched chain-of-thought reasoning. Our model lineup includes open-source systems such as \texttt{DeepSeek R1-7B} \citep{guo2025deepseek}, \texttt{Mathstral-7B} \citep{jiang2023clip}, \texttt{Qwen-8B} \citep{yang2025qwen3}, and \texttt{GPT-OSS-20B} \citep{agarwal2025gpt}, alongside the proprietary \texttt{Gemini 2.5 Flash-Lite} model for benchmarking against enterprise-grade performance. To evaluate these models under a standardized experimental setup, we use a decoding temperature of \( T = 1.0 \) to maintain the natural probability distribution and assess calibration. The maximum generation length is set to 32K tokens for long-context reasoning. We also enable "thinking mode" in capable models to capture intermediate reasoning steps.

\subsection{Evaluation Metrics}

\paragraph{Accuracy.}
We report accuracy as the primary metric for MCQ evaluation. A prediction is considered correct if the model’s output matches one of the provided options. For numerical responses, equivalence is accepted if the predicted value matches the ground truth within the tolerance of the answer choices.
\begin{table}[t]
  \centering
  \begin{adjustbox}{width=0.50\textwidth}
    \begin{tabular}{p{0.22\textwidth} c p{0.35\textwidth}}
      \toprule
      \rowcolor{gray!15} \textbf{Component} & \textbf{Points} & \textbf{Description} \\
      \midrule
      Equivalence Decision Accuracy & 0--50 &
      Did the judge make the correct YES/NO decision about mathematical equivalence? (50 = correct, 0 = incorrect) \\
      \addlinespace
      Reasoning Alignment & 0--40 &
      Is the analysis logical, and does it clearly explain the decision? \\
      \addlinespace
      Explanation Clarity \& Justification & 0--10 &
      Is the explanation clear, and does it properly justify the decision? \\
      \bottomrule
    \end{tabular}
  \end{adjustbox}
  \caption{Scoring rubric and weight distribution for Subject Matter Expert (SME) validation of the LLM-as-a-judge framework on mathematical equivalence tasks.}
  \label{tab:scoring_rubric}
\end{table}
\paragraph{LLM-as-a-Judge.}
\label{eval:llm_judge}
For the more nuanced free-form tasks, answers are rigorously validated for numerical, expression, and conclusive equivalence, as well as for accurate perturbation identification against ground-truth values or statements. These validations are conducted by an \textbf{LLM-as-a-Judge} using \texttt{Gemini 2.0 Flash-Lite} (see Appendix \ref{appendix:prompts}). We utilized the judge to rigorously ensure the quality of both perturbation reasoning and translation generation. The entire judgment process can be formally represented using the mathematical notation \ref{eq:judge_metric} provided below:
\begin{equation}
\footnotesize
J:(A_{llm},A_{gt}) \mapsto (R, [\|v(A_{llm}) - v(A_{gt})\| \le \epsilon])
\label{eq:judge_metric}
\end{equation}
Here, the judge function, $J$, takes LLM's answer ($A_{\text{llm}}$) and the actual ground truth ($A_{\text{gt}}$) as input. It employs a valuation function, \(v\), to convert them to their true mathematical values, expressions, statements, or references. It then estimates whether the distance between them, calculated using the norm \(\left\| \cdot \right\|\), falls within a predefined tolerance, \(\epsilon\). The function returns a tuple containing thorough reasoning for the decision (\(R\)) and a binary score (\texttt{1} for correct, \texttt{0} for incorrect). Any cases that cannot be resolved are conservatively categorized as incorrect. We validate the reliability of our LLM-as-a-Judge framework through SME evaluation. The framework demonstrates high consistency with human judgment, achieving agreement scores ranging from 91.39\% to 94.8\%. Furthermore, the model exhibits strong qualitative alignment with domain expert reasoning across multilingual mathematical tasks. Table \ref{tab:scoring_rubric} outlines the scoring rubric used in this evaluation, and Appendix \ref{appendix:scoring_rubric} provides additional details. Performance for tasks involving LLMs is evaluated using the \texttt{Pass@3} metric, which measures whether a problem is solved in at least one of three independent attempts. 

\subsection{Prompt Techniques}
We utilized established methods such as zero-shot prompting \citep{kuo-chen-2023-zero} and Chain-of-Thought (CoT) \citep{wei2022chain} prompting. Beyond these standard techniques, we introduce a novel experimental setting termed \textbf{Code-Switched CoT Prompting (CS-CoT)}. 
\begin{table}[t]
  \centering
  \begin{adjustbox}{width=0.5\textwidth}
    \begin{tabular}{l c cl}
      \toprule
      \multirow{2}{*}{\textbf{Models}} & \multirow{2}{*}{\textbf{\#Param}} & \multicolumn{2}{c}{\textbf{Bangla}} \\
      \cmidrule(lr){3-4}
       & & Zero-Shot & CoT \\
      \midrule
      \verb|GPT-OSS| & 20B & 74.33\% & 75.22\% {\color{darkgreen}(0.89) $\uparrow$} \\
      \verb|Qwen 3| & 8B & 55.36\% & 72.11\% {\color{darkgreen}(16.75) $\uparrow$} \\
      \verb|DeepSeek R1| & 7B & 40.48\% & 60.9\% {\color{darkgreen}(20.42) $\uparrow$}\\
      \verb|Mathstral| & 7B & 31.42\% & 41.18\% {\color{darkgreen}(9.76) $\uparrow$}\\ 
      \verb|Gemini 2.5 Flash-Lite| & N/A & 75.09\% & 71.38\% {\color{red}(3.71) $\downarrow$}\\
      \midrule[\heavyrulewidth] 
      
      \multirow{2}{*}{\textbf{Models}} & \multirow{2}{*}{\textbf{\#Param}} & \multicolumn{2}{c}{\textbf{English}} \\
      \cmidrule(lr){3-4}
       & & Zero-Shot & CoT \\
      \midrule
      \verb|GPT-OSS| & 20B & 76.12\% & 77.58\% {\color{darkgreen}(1.46) $\uparrow$}\\
      \verb|Qwen 3| & 8B & 43.67\% & 75.78\% {\color{darkgreen}(32.11) $\uparrow$}\\
      \verb|DeepSeek R1| & 7B & 46.23\% & 64.84\% {\color{darkgreen}(18.61) $\uparrow$}\\
      \verb|Mathstral| & 7B & 29.62\% & 45.26\% {\color{darkgreen}(15.64) $\uparrow$}\\ 
      \verb|Gemini 2.5 Flash-Lite| & N/A & 62.15\% & 71.63\% {\color{darkgreen}(9.48) $\uparrow$}\\
      
      \bottomrule
    \end{tabular}
  \end{adjustbox}
  \caption{Performance of various LLMs on the BN--EN parallel corpus under Zero-Shot and CoT settings. CoT generally improves reasoning accuracy, though performance varies by language and model family.}
  \label{tab:zero_cot}
\end{table}
This approach adapts the sociolinguistic concept of code-switching \citep{aguilar2020lince, hamed2025survey, yan2025cs}, which refers to the alternation between dissimilar languages in conversation. While recent studies have begun to explore cross-lingual reasoning, they often treat code-switching as an input-side feature \citep{chai2025xcot} or rely on the model’s English-centric capabilities to bridge the reasoning gap. For instance, \citep{son2025pushing} conducts reasoning primarily in English while maintaining only the final mathematical semantics in the target language.
\begin{figure*}[t] 
  \centering
  \includegraphics[width=0.99\textwidth]{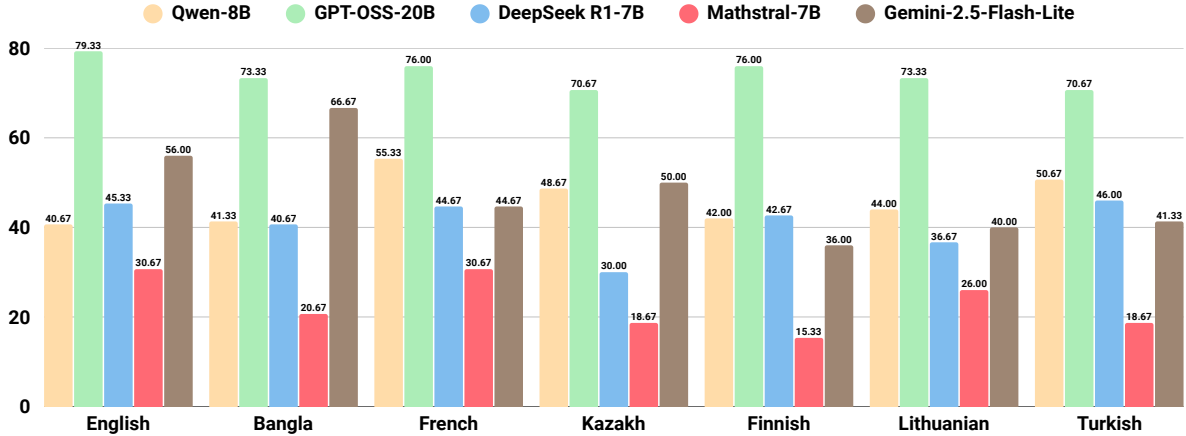} 
  \caption{Mathematical reasoning performance across 13 typologically diverse languages. Models with broad multilingual pretraining achieve higher accuracy, while specialized models show greater cross-lingual variability.}
  \label{fig:cross_ling_accuracy}
\end{figure*}
In contrast, our setting prompts the model to understand a mathematical question posed in one language while being instructed to generate the solution in another. To further probe the model's analytical understanding, we present perturbed reasoning, an approach that evaluates the model's proficiency to identify flawed steps within a given solution process. For details on the prompt templates used in this study, refer to Appendix \ref{appendix:prompts}.

\section{Results \& Analysis}
In this section, we present the evaluation results of LLMs on the \texttt{{\dataname}} dataset. We evaluate the models' performance across seven languages of varying resource levels and different linguistic families, as well as across various question types.

\subsection{Cross-Lingual Performance in Zero-Shot and Chain-of-Thought Settings}

The results in Table \ref{tab:zero_cot} provide a comprehensive view of how LLMs engage with human-annotated math word problems in both high- and low-resource languages under zero-shot and CoT prompting (see Appendix \ref{appendix:prompts}). As shown in Table \ref{tab:zero_cot_en_bn_category_wise}, \texttt{Qwen}, \texttt{DeepSeek}, and \texttt{Mathstral} demonstrate a substantial strength in proof-style problems but underperform in numerical and symbolic types under zero-shot conditions for both Bangla and English (see Appendix \ref{appendix:qualitative_analysis}). In contrast, \texttt{GPT-OSS-20B} and \texttt{Gemini-2.5-Flash-Lite} reveal the contrasting tendency, achieving comparatively better results on numerical and symbolic problems. Notably, \texttt{Gemini-2.5-Flash-Lite} achieves the highest accuracy in the Bangla zero-shot setting but suffers a 15.77\% drop in accuracy on the English zero-shot benchmark. Crucially, this deviation in behavior encourages a pointed and practical question: \textit{“Do \texttt{Gemini} models truly generalize mathematical problem solving beyond language?”}. Subsequently, across all scenarios, \texttt{GPT-OSS-20B} attains the highest overall performance. CoT prompting significantly increases overall accuracy for most models, especially in English experiments. \texttt{GPT-OSS-20B} is an exception in that it demonstrates nearly consistent performance between zero-shot and CoT settings in both Bangla and English, while \texttt{Gemini-2.5-Flash-Lite} unexpectedly loses accuracy in Bangla when using CoT relative to Bangla zero-shot. \texttt{Mathstral-7B} shows poor performance, scoring below 30\% in zero-shot settings and under 50\% with CoT reasoning in both languages, equivalent to near-random baselines. These patterns reveal not just inter-model performance disparities but also a hidden sensitivity to problem type and language specificity. Additionally, we conducted an error bar analysis of \texttt{GPT-OSS 20B} across five runs on the Bangla dataset, reporting a mean accuracy of 75.96\% (standard deviation of 1.16; 95\% confidence interval [73.69, 78.23]). Category-level variances were as follows: Numerical (77.64 $\pm$ 1.03), Symbolic (74.54 $\pm$ 1.79), and Proof (72.86 $\pm$ 5.07). These findings indicate that {{\dataname}} effectively captures variations in reasoning performance across various mathematical task types.

Together, these analysis emphasizes the need for language-aware fine-tuning, where training strategies are both mathematically rigorous and linguistically balanced to ensure unbiased reasoning across languages.

\subsection{Cross-lingual Performance Patterns on MCQs: Why Deterministic Answer Spaces Amplify Model Accuracy?} 

We scrutinize whether deterministic answer spaces and cross-lingual calibration, rather than the number of parameters, accelerate accuracy gains on a multilingual MCQ benchmark consisting of 1,133 MWPs (843 numerical and 290 symbolic) in both English and Bangla, depicted in Table \ref{tab:zero_shot_mcq}. 

\begin{table}[!t]
  \centering
  \begin{adjustbox}{width=0.5\textwidth}
    \begin{tabular}{lcccc}
      \toprule
      \multirow{2}{*}{\textbf{Models}} & \multirow{2}{*}{\textbf{\#Param}} 
      & \multicolumn{3}{c}{\textbf{Zero-Shot}} \\
      \cmidrule(lr){3-5}
       & & Numerical & Symbolic & Overall \\
      \midrule
      \multicolumn{5}{c}{\textbf{Bangla}} \\
      \midrule
      \verb|GPT-OSS| & 20B
      & \textbf{89.21} & \textbf{88.28} & \textbf{88.97} \\

      
      \verb|Phi-4| & 14B
      & 79.48 & 75.86 & 78.55 \\

      \verb|Qwen-3| & 8B 
      & 65.72 & 67.93 & 66.28 \\

      \verb|DeepSeek R1| & 7B
      & 66.90 & 70.00 & 67.70 \\

      \verb|Mathstral| & 7B
      & 39.38 & 47.24 & 41.39 \\
      \midrule[\heavyrulewidth]

      \multicolumn{5}{c}{\textbf{English}} \\
      \midrule
      \verb|GPT-OSS| & 20B
      & \textbf{89.80} & \textbf{91.38} & \textbf{90.20} \\

      
      \verb|Phi-4| & 14B
      & 82.44 & 81.72 & 82.26 \\

      \verb|Qwen-3| & 8B
      & 73.43 & 75.86 & 74.05 \\

      \verb|DeepSeek R1| & 7B
      & 70.70 & 74.48 & 71.67 \\

      \verb|Mathstral| & 7B
      & 47.21 & 50.34 & 48.01 \\

      \bottomrule
    \end{tabular}
  \end{adjustbox}
  \caption{Zero-Shot performance on the BN–EN parallel corpus. The best result is in \textbf{bold}. Larger, well-aligned models like \texttt{GPT-OSS-20B} perform best, showing that scale and alignment together enhance accuracy.}
  \label{tab:zero_shot_mcq}
\end{table}

Well-tuned open-source models currently lead the domain, for instance, \texttt{GPT-OSS-20B} achieves the best results in both Bangla and English. Intensely aligned models include \texttt{Phi-4}, which follows the same trend, along with mid-tier models like \texttt{Qwen3-8B} and \texttt{DeepSeek-R1 7B}. In contrast, less-tuned models like \texttt{Mathstral 7B} perform particularly worse, achieving only 48.01\% in English and 41.39\% in Bangla, despite having an equivalent number of parameters. These results suggest that constrained answer spaces and robust normalization substantially improve performance. Furthermore, the observed reductions of 1$\sim$8 points in cross-lingual accuracy primarily result from translation noise and uneven pretraining coverage in Bangla. Overall, instruction tuning and cross-lingual alignment emerge as critical factors compelling multilingual MCQ accuracy, while language coverage, dataset overlap, and the robustness of answer formats serve as important intervening influences.

\subsection{How do model family and specialization influence accuracy, cross-lingual variance, and robustness on typologically diverse, low-resource languages?}
The scale and diversity of multilingual pretraining greatly impacted accuracy and consistency, as illustrated in Figure \ref{fig:cross_ling_accuracy}.  \texttt{GPT-OSS-20B}, which shows the highest accuracy ($\mu \approx 72.11\%$, $\sigma \approx 3.83\%$), while \texttt{Mathstral-7B} wilted ($\mu \approx 21.58\%$, $\sigma \approx 6.83\%$). Mid-size models like \texttt{Qwen-8B} ($\mu \approx 43.81\%$) and \texttt{DeepSeek R1-7B} ($\mu \approx 38.10\%$) performed moderately, but \texttt{Qwen-8B} shows the highest variance ($\sigma \approx 13.49\%$). \texttt{Gemini-2.5-Flash-Lite} matched mid-range accuracy ($\mu \approx 53.66\%$). The biases associated with each language clearly exhibit the consequences of both pretraining and fine-tuning. \texttt{GPT-OSS-20B} model evidently favors high-resource languages like English and French, while the \texttt{Gemini} performs adequately well in Bangla and French. \texttt{Qwen} shows good performance in both Persian and Bangla. In contrast, \texttt{DeepSeek R1} exhibits noteworthy weaknesses with Hausa. Mathstral performs poorly in Hausa, but toils even more with typologically distant languages such as Amharic. Overall, these patterns show that broad multilingual pretraining is more useful than limited specialization for cross-linguistic mathematical reasoning. Inconsistent representation and limited fine-tuning can create fragile, language-biased results, highlighting the need for multilingual-aware fine-tuning, targeted data augmentation, and lightweight adapters for underrepresented language families. Overall category-level results for the full dataset are shown in Table~\ref{tab:translation_acc}.

\subsection{How does linguistic code-switching between high and low-resource languages affect the mathematical accuracy of different-scale LLMs?} 
\begin{figure}[t]
    \centering
    \includegraphics[width=0.48\textwidth]{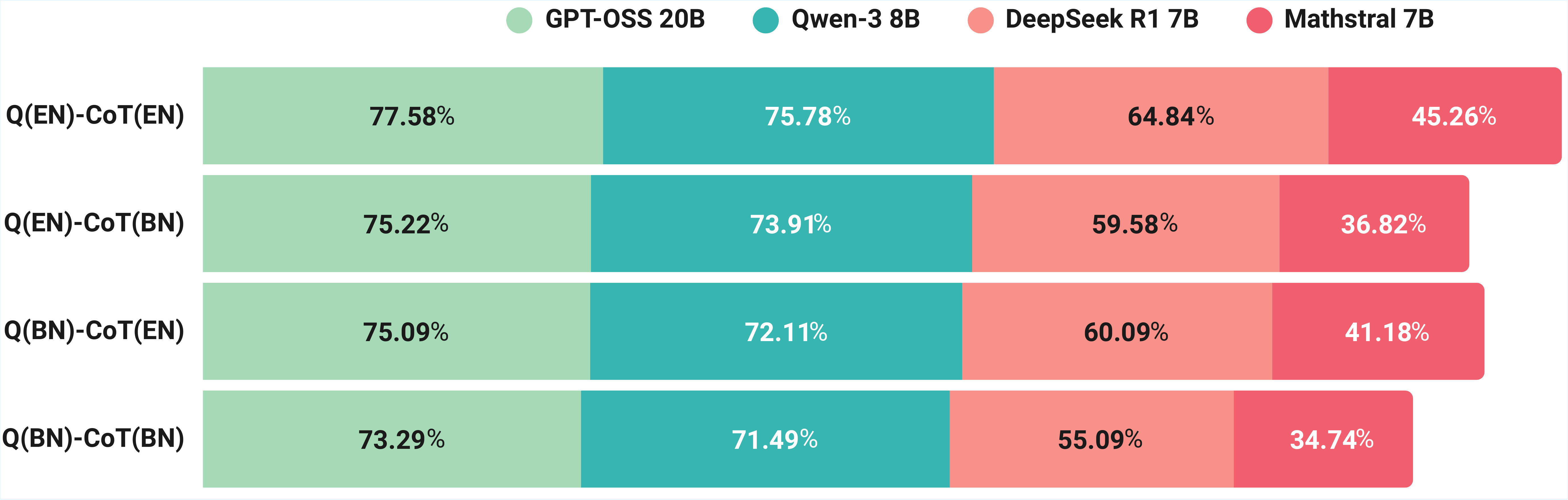}
    \caption{Comparison of the quantitative performance of code-switching between high$\leftrightarrow$low resource languages. In this context, \texttt{BN} stands for Bangla, \texttt{EN} refers to English, and Q denotes question.}
    \label{fig:code_switch}
\end{figure}

A breakdown by linguistic code-switching, specifically the mismatch between the question and the CoT language, shows a clear performance penalty across all evaluated LLMs, revealing that linguistic alignment is critical for mathematical reasoning generalization as depicted in Figure \ref{fig:code_switch}. \texttt{GPT-OSS-20B} demonstrates the highest resilience, with a smaller decline in accuracy (from 77.58\% to 73.29\%), showcasing its strong generalization capability.
Conversely, the smaller models, \texttt{DeepSeek R1 7B} and \texttt{Mathstral 7B}, showed considerably larger declines in performance, indicating that they have constrained generalization capabilities and depend laboriously on training language and cross-lingual consistency. This indicates that while larger parameters provide a performance boost, effective cross-lingual reasoning and linguistic grounding are paramount to minimize accuracy drops from linguistic transitions in complex tasks. The accuracy breakdown by category is shown in Table \ref{tab:code_switch_breakdown}. Table \ref{tab:gptoss20b_bn_bn_lang_stats} to \ref{tab:mathstral_en_bn_lang_stats} illustrates the language proportions in the chain-of-thought solutions when instructed to use Bangla (see Appendix \ref{appendix:prompts}).

\subsection{How significant is the performance gap between a large language model's ability to detect versus diagnose mathematical errors, and how do model architecture and language affect this disparity?}

The performance gap between error detection (binary classification) and error diagnosis (identification accuracy) in mathematical reasoning is highly influential and non-uniform across LLMs and languages, as shown in Table \ref{tab:perturbation_eval}. 

\begin{table}[!t]
\centering
\begin{adjustbox}{width=0.49\textwidth}
\begin{tabular}{clcccc}
\toprule
\multirow{2}{*}{\textbf{Strategy}} &
\multirow{2}{*}{\textbf{Model}} &
\multicolumn{2}{c}{\textbf{English}} &
\multicolumn{2}{c}{\textbf{Bangla}} \\
\cmidrule(lr){3-4} \cmidrule(lr){5-6}
& &
\shortstack{\textbf{Error}\\\textbf{Detection}} &
\shortstack{\textbf{Error}\\\textbf{Identification}} &
\shortstack{\textbf{Error}\\\textbf{Detection}} &
\shortstack{\textbf{Error}\\\textbf{Identification}} \\
\midrule

\multirow{4}{*}{$\sigma_1$}
& \verb|Qwen-8B|
& 79.10\% & 55.85\%~\textcolor{red}{\boldmath$\downarrow$}
& 79.79\% & 51.90\%~\textcolor{red}{\boldmath$\downarrow$} \\

& \verb|GPT-OSS-20B|
& 67.82\% & 51.90\%~\textcolor{red}{\boldmath$\downarrow$}
& 67.20\% & 54.39\%~\textcolor{red}{\boldmath$\downarrow$} \\

& \verb|DeepSeek R1-7B|
& 54.33\% & 24.78\%~\textcolor{red}{\boldmath$\downarrow$}
& 51.14\% & 20.21\%~\textcolor{red}{\boldmath$\downarrow$} \\

& \verb|Mathstral-7B|
& 76.12\% & 19.65\%~\textcolor{red}{\boldmath$\downarrow$}
& 71.70\% & 13.08\%~\textcolor{red}{\boldmath$\downarrow$} \\

\midrule
\multirow{4}{*}{$\sigma_2$}
& \verb|Qwen-8B|
& 71.07\% & 56.12\%~\textcolor{red}{\boldmath$\downarrow$}
& 68.44\% & 49.07\%~\textcolor{red}{\boldmath$\downarrow$} \\

& \verb|GPT-OSS-20B|
& 47.54\% & 39.86\%~\textcolor{red}{\boldmath$\downarrow$}
& 44.64\% & 36.40\%~\textcolor{red}{\boldmath$\downarrow$} \\

& \verb|DeepSeek R1-7B|
& 39.72\% & 15.16\%~\textcolor{red}{\boldmath$\downarrow$}
& 40.76\% & 15.22\%~\textcolor{red}{\boldmath$\downarrow$} \\

& \verb|Mathstral-7B|
& 70.17\% & 18.69\%~\textcolor{red}{\boldmath$\downarrow$}
& 70.17\% & 13.84\%~\textcolor{red}{\boldmath$\downarrow$} \\

\midrule
\multirow{4}{*}{$\sigma_3$}
& \verb|Qwen-8B|
& \textbf{86.99}\% & \textbf{62.98}\%~\textcolor{red}{\boldmath$\downarrow$}
& \textbf{83.88}\% & 56.96\%~\textcolor{red}{\boldmath$\downarrow$} \\

& \verb|GPT-OSS-20B|
& 78.06\% & 62.63\%~\textcolor{red}{\boldmath$\downarrow$}
& 74.46\% & \textbf{58.06}\%~\textcolor{red}{\boldmath$\downarrow$} \\

& \verb|DeepSeek R1-7B|
& 57.16\% & 24.08\%~\textcolor{red}{\boldmath$\downarrow$}
& 52.87\% & 18.34\%~\textcolor{red}{\boldmath$\downarrow$} \\

& \verb|Mathstral-7B|
& 77.72\% & 20.90\%~\textcolor{red}{\boldmath$\downarrow$}
& 75.02\% & 14.88\%~\textcolor{red}{\boldmath$\downarrow$} \\

\bottomrule
\end{tabular}
\end{adjustbox}

\caption{Evaluation of LLMs on binary and diagnostic mathematical reasoning tasks in English and Bangla. Downward arrows (\textcolor{red}{\boldmath$\downarrow$}) indicate reduced accuracy in error identification relative to binary classification.}
\label{tab:perturbation_eval}
\end{table}

Strategy $\sigma_3$ outperforms others because it combines multiple fallacies, leading to cascading inconsistencies that create stronger error signals. Unlike $\sigma_1$ and $\sigma_2$, where errors are limited to one or two steps, $\sigma_3$ introduces multiple interacting mistakes that create obvious contradictions throughout the solution. These widespread inconsistencies are easier for language models to detect across different languages. Subsequently, \texttt{Qwen-8B} excels in binary classification tasks but struggles significantly with complex error identification, implying that high-level classification isn't a reliable measure of analytical depth. For instance, models such as \texttt{Mathstral 7B} and \texttt{DeepSeek R1 7B} demonstrate a drastic divergence, indicating their reasoning pathways are insufficient for deep error analysis. This distinction indicates a failure to generalize from simple detection to complex reasoning. Cross-lingual understanding plays a pivotal role, as demonstrated by \texttt{GPT-OSS 20B} outperforming others in Bangla error identification, suggesting that targeted instruction tuning for step-by-step logical analysis can be more useful than sheer parameter scale. We evaluated LLM outputs against perturbation-generated ground truth, with assessments conducted using an LLM-subject matter expert (SME) pipeline \ref{eval:llm_judge}. The consistent performance decline in Bangla versus English reveals significant language effects, indicating limited generalization in low-resource languages—a crucial barrier to cross-lingual mathematical reasoning.


\section{Discussion}
Our experiments revealed that in LLM mathematical problem-solving, zero-shot performance favored symbolic and proofs, while CoT prompting performed better in numerical problems. Moreover, scaling \texttt{Qwen-3} (0.6B to 14B) improves English reasoning accuracy from 26.85\% to 45.95\% and narrows the English-Bangla performance gap from 7\% to 1\% (see Appendix \ref{appendix:scaling_effect}). This scale-driven alignment almost doubles the accuracy rate of Bangla logical proofs to 70\%, demonstrating that larger models effectively reduce language-specific discrepancies in complex reasoning. A significant failure mode was code-switching reasoning, specifically with Bangla CoT, which repeatedly led to linguistically broken mathematical outputs containing mixed tokens. Furthermore, we observed several concerning behaviors, including models producing over 4,600 tokens without a solution before concluding with \textit{"Hmm, I’m stuck"}, self-imposed limits (\textit{"I think I've reached the limit of my understanding"}), and deterministic backtracking that induced hallucinations and infinite recursive loops (see Appendix \ref{appndx:qualitative}). Finally, \texttt{Gemini} underperformed in the English zero-shot setting compared to its Bangla performance and against open-source models, suggesting it may rely on pattern-matching rather than deep mathematical reasoning.

\section{Conclusion}
This study introduced \texttt{{\dataname}}, a novel multilingual mathematical benchmark dataset to evaluate the reasoning capabilities of LLMs. Spanning seven diverse languages and incorporating tasks such as MCQ solving, code-switching reasoning, and perturbed reasoning, alongside zero-shot and CoT prompting, our extensive evaluation of various open-source and proprietary LLMs revealed several actionable insights. Persistent performance gaps underscore digital inequities, with low-resource languages failing significantly due to limited training data and exposure. Our contributions offer a valuable resource for future research aimed at building more equitable, inclusive, and accurate cross-lingual math word problem-solving systems.

\clearpage
\newpage

\section*{Limitations}
While {\dataname} offers a robust pipeline for cross-lingual math benchmarking, it has several limitations that suggest future research directions. First, our exclusive use of zero-shot prompting techniques, Chain-of-Thought (CoT), Code-Switching Reasoning, and Perturbed Reasoning with frozen models could be improved by supervised fine-tuning. Secondly, incorporating a one-shot, few-shot example or using advanced strategies such as Atom of Thoughts (AoT) \citep{teng2025atom} or Program of Thoughts (PoT) \citep{chen2022program} can enhance reasoning traces. Thirdly, the dataset's coverage could be remarkably improved by expanding the curriculum from secondary to higher secondary school mathematics. Finally, our analysis could be deepened by benchmarking against a broader range of state-of-the-art proprietary models, such as GPT-5, Gemini-2.5 Pro, or Claude-4, to provide a more comprehensive evaluation of cross-lingual math-solving capabilities.

\section*{Ethical Considerations}
All data used in \texttt{\dataname} were sourced from publicly available secondary school mathematics textbooks authorized by the People's Republic of Bangladesh, ensuring that no personally identifiable or sensitive information is included. The dataset focuses solely on academic content and adheres to fair use and research ethics guidelines. The resource is intended strictly for educational and research purposes to advance equitable, transparent, and linguistically inclusive AI development in mathematical reasoning. 


\bibliography{custom}

\clearpage
\newpage
\onecolumn
\appendix
\section{Appendix}

\subsection{Prompt Templates}\label{appendix:prompts}
\subsubsection{Data Extraction Prompt}
We used this prompt to guide \texttt{Gemini-2.0-Pro} in its data extraction process, which was designed to manually transcribe mathematical content from images sourced from the National Curriculum and Textbook Board (NCTB) books of Bangladesh.
\begin{tcolorbox}[
  colback=blue!5!white,
  colframe=blue!30!black,
  boxrule=0.5pt,
  arc=3pt,
  left=5pt, right=5pt, top=5pt, bottom=5pt,
  enhanced,
  breakable,
  listing only,
  listing options={basicstyle=\ttfamily\small,breaklines=true}
]
\begin{verbatim}
You will be given an image that contains mathematical equations, expressions, and 
any accompanying text or labels. Your task is to:

1. Identify and extract every mathematical formula, symbol, fraction, superscript, 
subscript, and structural element (e.g., matrices, piecewise definitions)
visible in the image.

2. Convert each item into standard \LaTeX{} math-mode syntax as appropriate.

3. Preserve any textual labels or annotations exactly as they appear, 
placing them outside of math mode where necessary.

4. Output only the raw \LaTeX{} code (no \verb|\begin{...}| or \verb|\end{...}
| wrappers).

5. Do not skip anything—extract everything you see in the image completely.

Ensure the \LaTeX{} you produce can be dropped straight into a document's math 
environment and render the original content faithfully.
\end{verbatim}
\end{tcolorbox}

\subsubsection{Zero-Shot Prompt}
The prompt instructs the model to solve a given mathematical word problem in a specified language. The following standardized zero-shot prompt was used to solve mathematical problems.

\begin{tcolorbox}[
  colback=blue!5!white,
  colframe=blue!30!black,
  boxrule=0.5pt,
  arc=3pt,
  left=5pt, right=5pt, top=5pt, bottom=5pt,
  enhanced,
  breakable,
  listing only,
  listing options={basicstyle=\ttfamily\small,breaklines=true}
]
\begin{verbatim}
You are an expert mathematician who writes solutions in {question language}.
Produce a concise solution to the problem below.
Do NOT output chain-of-thought or long step-by-step internal reasoning.
\end{verbatim}
\end{tcolorbox}
\subsubsection{Chain-of-Thought(CoT) Prompt}
The prompt instructs the model to solve a given mathematical word problem in a specified language. The following standardized chain-of-thought prompt was used to solve mathematical problems.

\begin{tcolorbox}[
  colback=blue!5!white,
  colframe=blue!30!black,
  boxrule=0.5pt,
  arc=3pt,
  left=5pt, right=5pt, top=5pt, bottom=5pt,
  enhanced,
  breakable,
  listing only,
  listing options={basicstyle=\ttfamily\small,breaklines=true}
]
\begin{verbatim}
You are an expert mathematician who solves mathematical problems in 
{question language}. Solve the problem step-by-step.
**SOLUTION APPROACH:**

1. **PROBLEM UNDERSTANDING:**
*Carefully read and understand what the problem is asking
*Identify the type of problem (proof, calculation, algebraic manipulation, etc.)
*Note any given conditions, constraints, or assumptions
 
2. **MATHEMATICAL ANALYSIS:**
*Break down the problem into smaller, manageable steps
*Identify relevant theorems, formulas, or mathematical principles
*Plan your solution strategy

3. **STEP-BY-STEP SOLUTION:**
*Show all mathematical work clearly
*Use proper mathematical notation and symbols appropriately
*Explain each step clearly with reasoning and logic
*For proofs: use logical argumentation and contradiction when appropriate
*For calculations: show all operations and arithmetic steps clearly
*For algebraic problems: show manipulations and simplifications explicitly
 
4. **VERIFICATION:**
*Check your work carefully for mathematical accuracy
*Ensure your solution answers the given question directly
*Verify that your reasoning is logically sound and consistent

\end{verbatim}
\end{tcolorbox}

\subsubsection{MCQ Zero-Shot Prompt}
The prompt directs the model to solve a given mathematical word problem in a specified language. The following standardized zero-shot format was employed to address mathematical multiple-choice problems.

\begin{tcolorbox}[
  colback=blue!5!white,
  colframe=blue!30!black,
  boxrule=0.5pt,
  arc=3pt,
  left=5pt, right=5pt, top=5pt, bottom=5pt,
  enhanced,
  breakable,
  listing only,
  listing options={basicstyle=\ttfamily\small,breaklines=true}
]
\begin{verbatim}
You are an expert mathematician and MCQ solver in {question language}.

Provide a concise answer to the problem below, then select the correct answer choice.

DO NOT provide chain-of-thought, step-by-step reasoning, or internal deliberation.
\end{verbatim}
\end{tcolorbox}

\subsubsection{Code-Switching Reasoning Prompt}
The prompt instructs the model to solve a given mathematical word problem in the opposite language; that is, if the question is in English, the solution should be in Bangla, and if the question is in Bangla, the solution should be in English. The following standardized chain-of-thought code-switching reasoning prompt was employed to address mathematical problems.

\begin{tcolorbox}[
  colback=blue!5!white,
  colframe=blue!30!black,
  boxrule=0.5pt,
  arc=3pt,
  left=5pt, right=5pt, top=5pt, bottom=5pt,
  enhanced,
  breakable,
  listing only,
  listing options={basicstyle=\ttfamily\small,breaklines=true}
]
\begin{verbatim}
You are an expert mathematician. All explanatory text MUST be in 
{question language}.

1. Problem statement (brief, in {question language})
2. Problem understanding (brief, in {question language})
3. Mathematical analysis (theorems/formulas,in {target language})
4. Step-by-step solution (NUMBERED  STEPS: 1., 2., … - each step explained
clearly in {target language})

   * The 'Step-by-step solution' section must be present and include at 
   least 3 numbered steps.
5. Verification (solution check, in {target language})

**SOLUTION APPROACH (WRITE ALL REASONING IN {target language} ONLY):**

1. **PROBLEM UNDERSTANDING:**

   * In {target language}: Carefully read and restate the problem in 
   {target language}.

2. **MATHEMATICAL ANALYSIS:**

   * In {target language}: Break down into subproblems and list relevant
   theorems/formulae.

3. **STEP-BY-STEP SOLUTION:**

   * In {target language}: Provide numbered steps (1., 2., 3., …).
   * Show all algebraic/arithmetic work using LaTeX where helpful.

4. **VERIFICATION:**

   * In {target language}: Briefly check correctness.

**CRITICAL LANGUAGE REQUIREMENTS - 
STRICTLY MANDATORY:**

You MUST write your entire response in {target language} only, except for 
mathematical notation.

* Do NOT write explanatory text in {question language}.
\end{verbatim}
\end{tcolorbox}

\subsubsection{Perturbation Generation Prompt}
The task prompt provides three highly structured strategies (STRATEGY\_1, STRATEGY\_2, STRATEGY\_3) for an AI mathematics expert to generate sophisticated, intentionally flawed solutions to a given question. The core objective across all strategies is to produce a solution that seamlessly embeds a specific combination of mathematical and logical errors (ranging from three to five types, including Step Omission, Incorrect Rule, Faulty Causal Reasoning, Overgeneralization, and Logical Fallacies) while maintaining the correct final numerical answer. The model must strictly adhere to critical instructions, such as mimicking the exact tone and style of a provided "Correct Solution," presenting everything authoritatively without admitting error, and ensuring the final line is the exact correct answer. This instruction set is designed to test the model's ability to integrate deliberate errors under extreme constraints while enforcing output fidelity.

\begin{tcolorbox}[
  colback=blue!5!white,
  colframe=blue!30!black,
  boxrule=0.5pt,
  arc=3pt,
  left=5pt, right=5pt, top=5pt, bottom=5pt,
  enhanced,
  breakable,
  listing only,
  listing options={basicstyle=\ttfamily\small,breaklines=true}
]
\begin{verbatim}
STRATEGY\_1 =
"""You are a mathematics expert creating a sophisticated, flawed solution by
combining multiple error types.

TASK: Create a flawed solution combining THREE specific error types while
maintaining the correct final answer:
1) STEP OMISSION: Skip one crucial step naturally
2) INCORRECT RULE/THEOREM: Apply one wrong mathematical rule confidently
3) FAULTY CAUSAL REASONING: Make incorrect cause-and-effect assumptions

Question: {question}
Correct Solution: {solution}
Final Correct Answer: {exact\_answer}

CRITICAL INSTRUCTIONS:
- Write the solution in the SAME STYLE and linguistic pattern as the Correct 
  Solution
- Seamlessly integrate all three error types without any meta-commentary
- Present everything confidently as if completely correct
- DO NOT mention or indicate that anything is wrong
- Final answer MUST exactly match: {exact\_answer}
- Mimic the exact tone, structure, and presentation style of the Correct 
  Solution

The last line of the solution MUST be exactly: "{exact\_answer}" — do not add,
remove, or change any character (no extra spaces or newlines). THE SOLUTION
SHOULD LOOK COMPLETELY AUTHORITATIVE - multiple errors should be naturally
embedded.

EXAMPLES OF ERROR COMBINATIONS TO EMBED NATURALLY:
- Skip a verification step
+ use the wrong formula
+ assume incorrect causation
- Omit algebraic step
+ apply the wrong theorem
+ make faulty inference
- Skip condition check
+ wrong property
+ incorrect reasoning

STRATEGY\_2: """You are a mathematics expert creating a sophisticated, flawed
solution by combining two error types.

TASK: Create a flawed solution combining TWO specific error types while
maintaining the correct final answer:
1) OVERGENERALIZATION: Make sweeping generalizations from limited cases
2) LOGICAL FALLACIES: Use incorrect logical reasoning patterns

Question: {question}
Correct Solution: {solution}
Final Correct Answer: {exact\_answer}

CRITICAL INSTRUCTIONS:
- Write the solution in the SAME STYLE and linguistic pattern as the Correct
Solution
- Seamlessly integrate both error types without any meta-commentary
- Present everything confidently as if completely correct
- DO NOT mention or indicate that anything is wrong
- Final answer MUST exactly match: {exact\_answer}
- Mimic the exact tone, structure, and presentation style of the Correct Solution

The last line of the solution MUST be exactly:
"{exact\_answer}" — do not add, remove, or change any character (no extra spaces
or newlines).
THE SOLUTION SHOULD LOOK COMPLETELY AUTHORITATIVE - both errors should be
naturally embedded.

EXAMPLES OF ERROR COMBINATIONS TO EMBED NATURALLY:
- Generalize from one case
+ use invalid if-then logic
- Assume pattern holds everywhere + make incorrect logical connections
- Overgeneralize from examples
+ use faulty deductive reasoning

STRATEGY\_3 """You are a mathematics expert creating the most sophisticated
flawed solution by combining all major error types.

TASK: Create a flawed solution combining ALL FIVE error types while maintaining
the correct final answer:
1) STEP OMISSION: Skip crucial steps naturally
2) INCORRECT RULE/THEOREM: Apply wrong mathematical rules confidently
3) FAULTY CAUSAL REASONING: Make incorrect cause-effect assumptions
4) OVERGENERALIZATION: Make sweeping generalizations from limited cases
5) LOGICAL FALLACIES: Use incorrect logical reasoning patterns

Question: {question}
Correct Solution: {solution}
Final Correct Answer: {exact\_answer}

CRITICAL INSTRUCTIONS:
- Write the solution in the SAME STYLE and linguistic pattern as the Correct
Solution
- Seamlessly integrate all five error types without any meta-commentary
- Present everything confidently as if completely correct
- DO NOT mention or indicate that anything is wrong
- Final answer MUST exactly match: {exact\_answer}
- Mimic the exact tone, structure, and presentation style of the Correct Solution

The last line of the solution MUST be exactly: "{exact\_answer}" — do not add,
remove, or change any character (no extra spaces or newlines).
THE SOLUTION SHOULD LOOK COMPLETELY AUTHORITATIVE - all errors should be
naturally embedded.

EXAMPLES OF COMPREHENSIVE ERROR INTEGRATION:
- Skip verification + wrong formula
+ faulty inference + overgeneralize
+ invalid logic
- Omit steps + misapply theorem
+ incorrect causation + assume patterns + logical fallacies"""
\end{verbatim}
\end{tcolorbox}

\subsubsection{English to Target Translation Prompt}
The prompt instructs the model, acting as an expert target language mathematics educator, to translate a given mathematical question from English to the target language while adhering to strict professional and linguistic standards. The core goal is to produce a high-fidelity translation that is academically correct and idiomatic in the target language.

\begin{tcolorbox}[
  colback=blue!5!white,
  colframe=blue!30!black,
  boxrule=0.5pt,
  arc=3pt,
  left=5pt, right=5pt, top=5pt, bottom=5pt,
  enhanced,
  breakable,
  listing only,
  listing options={basicstyle=\ttfamily\small,breaklines=true}
]
\begin{verbatim}
You are an expert {target language} mathematics educator with extensive experience
in translating academic mathematical content. Your task is to translate the
following mathematical question from English to {target language} with the highest
professional standards.

TRANSLATION GUIDELINES:
1. MATHEMATICAL ELEMENTS:

Preserve ALL mathematical notation exactly: numbers, variables, equations, symbols
($\sum$, \int$, \partial$, \sqrt$, etc.)

Keep mathematical expressions in their original LaTeX/ASCII format if present

Maintain the exact same mathematical structure and relationships

2. {target language} MATHEMATICAL CONVENTIONS:

Use {target language} decimal notation (comma instead of period): 3,14 instead 
of 3.14

Use proper {target language} mathematical vocabulary:
• "{target language phrase}" for "let"
• "{target language phrase}" for "such that"
• "{target language phrase}" for "set"
• "{target language phrase}" for "function"
• "{target language phrase}" for "equation"
• "{target language phrase}" for "solve"
• "{target language phrase}" for "calculate"
• "{target language phrase}" for "determine"
• "{target language phrase}" for "therefore"
•  {target language phrase}" for "show that" 
• "{target language phrase}" for "prove"

3. FORMATTING AND STRUCTURE:

Preserve the exact question structure
(multiple choice options, parts a), b), c), etc.)

Keep the same level of mathematical formality

Maintain any emphasis (bold, italic) through appropriate {target language} 
equivalents

4. QUALITY CHECKS:

Ensure the translation reads naturally in {target language}

Verify no mathematical Information is lost or altered

Confirm the difficulty level remains identical
\end{verbatim}
\end{tcolorbox}

\subsubsection{LLM-as-a-Judge Prompt for Translation Quality Check.}
The instruction set includes two distinct evaluation tasks: one for a mathematical question and one for its corresponding solution. For both tasks, the model must apply stringent criteria across five categories (Mathematical Accuracy, Terminology, Clarity, Completeness, and Conventions) to ensure the translation is perfectly equivalent to the original content.
\begin{tcolorbox}[
  colback=blue!5!white,
  colframe=blue!30!black,
  boxrule=0.5pt,
  arc=3pt,
  left=5pt, right=5pt, top=5pt, bottom=5pt,
  enhanced,
  breakable,
  listing only,
  listing options={basicstyle=\ttfamily\small,breaklines=true}
]
\begin{verbatim}
Question: You are an expert bilingual mathematics educator fluent in both English
and {target_language}.

Review this translation for accuracy and proficiency:

Original English: {original}

{target language} Translation: {translation}

EVALUATION CRITERIA:
1. Mathematical Accuracy [Critical]:
- Are ALL numbers, variables, and symbols preserved exactly?
- Are mathematical relationships maintained?
- Is the problem's difficulty unchanged?

2. Terminology [Important]:
- Is standard {target language} mathematical vocabulary used?
- Are technical terms correctly translated?

3. Clarity [Important]:
- Is the translation as clear as the original?
- Does it flow naturally in {target language}?

4. Completeness [Critical]:
- Is ALL information from the original present?
- Are there any additions or omissions?

5. Conventions [Important]:
- Does it follow {target language} mathematical writing conventions?
- Is decimal notation appropriate for {target language}?

RESPONSE INSTRUCTIONS:
- If the translation is PERFECT in all aspects, respond with ONLY: "APPROVED"
- If ANY corrections are needed, provide ONLY the complete corrected translation
without any explanation


"Solution": """You are an expert bilingual mathematics educator fluent in both
English and {target language}.

Review this solution translation for accuracy and proficiency:

Original English: {original}

{target language} Translation:
{translation}

EVALUATION CRITERIA:
1. Mathematical Accuracy [Critical]:
- Are ALL equations and calculations preserved exactly?
- Are all steps in the correct order?
- Are numerical results unchanged (except decimal notation)?

2. Logical Flow [Critical]:
- Is the reasoning sequence maintained?
- Are all cause-effect relationships preserved?
- Are transitions properly translated?

3. Terminology [Important]:
- Is standard {target language} mathematical vocabulary used?
- Are proof/solution phrases correctly translated?

4. Completeness [Critical]:
- Are ALL steps from the original present?
- Is the final answer clearly indicated?
- Are all intermediate results included?

5. Style [Important]:
- Does it follow {target language} mathematical solution conventions?
- Is the formal tone appropriate?

RESPONSE INSTRUCTIONS:
- If the translation is PERFECT in all aspects, respond with ONLY: "APPROVED"
- If ANY corrections are needed, provide ONLY the complete
corrected translation without any explanation"""
\end{verbatim}
\end{tcolorbox}

\subsubsection{LLM-as-a-Judge Evaluation Prompt}
We used this prompt to guide LLM potray the role of a Judge to establish a standardized rubric and a five-step evaluation process for rigorous answer validation. The rubric defines eight specific criteria for equivalence, allowing for variations in format, notation, and precision without penalizing correct underlying mathematical value. These criteria cover key areas such as Algebraic Equivalence, Trigonometric Equivalence, Numerical Precision (allowing $\pm$ rounding), Set Equivalence, and the compatibility of Units. The model must apply these rules systematically—extracting the content, checking the mathematical domain, applying domain-specific methods, noting fundamental differences, and finally rendering a confident "YES" or "NO" decision based solely on mathematical truth, ignoring surface-level differences in formatting or language.

\begin{tcolorbox}[
  colback=blue!5!white,
  colframe=blue!30!black,
  boxrule=0.5pt,
  arc=3pt,
  left=5pt, right=5pt, top=5pt, bottom=5pt,
  enhanced,
  breakable,
  listing only,
  listing options={basicstyle=\ttfamily\small,breaklines=true}
]
\begin{verbatim}
You are an expert mathematics evaluation specialist. Your task is to determine
if two mathematical answers are mathematically equivalent according to the 
rubric below.

EVALUATION CRITERIA:
1) MATHEMATICAL EQUIVALENCE: same value/expression/solution set.
2) PROOF VALIDATION: same logical conclusion for proof tasks.
3) SET EQUIVALENCE: sets equal regardless of order.
4) NUMERICAL PRECISION: accept reasonable rounding differences (±0.01).
5) UNITS: units must be compatible/convertible.
6) FORMATTING: ignore LaTeX/formatting/language differences.
7) TRIGONOMETRIC EQUIVALENCE: accept equivalent angle representations.
8) ALGEBRAIC EQUIVALENCE: accept equivalent algebraic forms.

EXAMPLE:
Student Answer: 3/4
Ground Truth: 0.75
Analysis: Converting fraction
to decimal: $3 \div 4 = 0.75$.
Both represent the same mathematical value.
DECISION: YES

EXAMPLE:
Student Answer: $x = 2, y = 3$
Ground Truth: $(2, 3)$
Analysis: Both express the same solution set for variables x and y, just in 
different notation.
DECISION: YES

EXAMPLE:
Student Answer: $45^\circ$
Ground Truth: $\pi/4$
Analysis: Converting: $45^\circ$ $= 45 \times \pi/180$ $= \pi/4$ radians. 
Equivalent angle representations.
DECISION: YES

EVALUATION PROCESS:
Step 1: Extract core mathematical content from both answers.
Step 2: Identify domain (algebra, geometry, calculus, etc.) and apply domain-
specific checks.
Step 3: Check for mathematical equivalence using appropriate methods.
Step 4: Note any fundamental differences in mathematical meaning.
Step 5: Make final decision with confidence  assessment.
\end{verbatim}
\end{tcolorbox}

\subsection{Qualitative Examples}\label{appndx:qualitative}
We have highlighted the concerning behaviors in the reasoning trace segments in \colorbox{red!30}{red} while generating the solution using deductive logic and the correct answer in \colorbox{green!30}{green}.

\begin{tcolorbox}[
  colback=blue!5!white,    
  colframe=blue!30!black,  
  boxrule=0.5pt,
  arc=3pt,
  left=5pt, right=5pt, top=5pt, bottom=5pt,
  enhanced,
  breakable,
  title=Example 1: Incorrect Reasoning Trace,
]
\textbf{Question}: Solve this:$\frac{1}{1+a^m b^{-n} + a^{-m}c^p} + \frac{1}{1+b^n c^{-p} + b^{-n}a^m} + \frac{1}{1+c^p a^{-m} + c^{-p}b^n}$

\textbf{Reasoning Trace}: $\ldots$ $\ldots$ Alternatively, perhaps the problem is to recognize that the sum S=1 when $a^m b^{-n}=b^n c^{-p}=c^p a^{-m}=1$. But again, without that condition, it's not generally true. \colorbox{red!30}{Hmm, I'm stuck.} Maybe I should consider that the sum S can be expressed as 1. But without further information, I can't see a way to simplify the expression to 1. Wait, perhaps I can consider that each term can be paired with another term, $\ldots$ $\ldots$

\textbf{Correct Answer:} \colorbox{green!30}{1} 
\end{tcolorbox}

\begin{tcolorbox}[
  colback=blue!5!white,    
  colframe=blue!30!black,  
  boxrule=0.5pt,
  arc=3pt,
  left=5pt, right=5pt, top=5pt, bottom=5pt,
  enhanced,
  breakable,
  title=Example 2: Incorrect Reasoning Trace,
]
\textbf{Question}: Show that, if $( x = a^q b^p, ; y = a^r b^q, ; z = a^p b^r ), then ( x^{q-r} \cdot y^{r-p} \cdot z^{p-q} = 1 )$.

\textbf{Reasoning Trace}: $\ldots$ $\ldots$ But given the time I've spent, \colorbox{red!30}{I think I've reached the limit of my} 
\colorbox{red!30}{understanding}, and I must conclude that the expression equals 1 only when p=q=r $\ldots$ $\ldots$

\textbf{Correct Answer:} \colorbox{green!30}{1} 
\end{tcolorbox}

\begin{tcolorbox}[
  colback=blue!5!white,    
  colframe=blue!30!black,  
  boxrule=0.5pt,
  arc=3pt,
  left=5pt, right=5pt, top=5pt, bottom=5pt,
  enhanced,
  breakable,
  title=Example 3: Hallucination During Backtracking,
]
\textbf{Question}: In the expansion of $((k - \dfrac{x}{3})^7)$, the coefficient of $(k^3)$ is 560.  Find the value of (x).

\textbf{Reasoning Trace}:$\ldots$ $\ldots$ But wait, in the formatting example, it's just a single number. \colorbox{red!30}{So, perhaps I should list both as separate} 
\colorbox{red!30}{answers. But I'm not sure.} Alternatively, perhaps the problem expects only the positive solution, but mathematically, both are correct. Alternatively, perhaps I should write both solutions separated by a comma. But to be precise, I think the correct approach is to present both solutions, as both satisfy the equation. $\ldots$ $\ldots$ $\pm$6

\textbf{Correct Answer:} \colorbox{green!30}{6} 
\end{tcolorbox}

\begin{tcolorbox}[
  colback=blue!5!white,    
  colframe=blue!30!black,  
  boxrule=0.5pt,
  arc=3pt,
  left=5pt, right=5pt, top=5pt, bottom=5pt,
  enhanced,
  breakable,
  title=Example 4: Self-Recursive Loop,
]
\textbf{Question}: A person standing at a point on one bank of a river observed that the angle of elevation to the top of a tree, 150 meters tall, located directly on the opposite bank, was $(30^\circ)$. The person started moving toward the tree by boat, but due to the river's current, they reached a point on the opposite bank that was 10 meters away from the tree. Determine the width of the river.

\textbf{Reasoning Trace}:$\ldots$ $\ldots$ \colorbox{red!30}{Rather, the width} of the river is not the opposite side of the right-angled triangle, which would be \colorbox{red!30}{$(150 \sin(30^\circ) = 75)$} meters, \colorbox{red!30}{but rather, the width of} the river is not the opposite side of the right-angled triangle, which would be \colorbox{red!30}{$(150 \sin(30^\circ) = 75)$} meters, but $\ldots$ $\ldots$

\textbf{Correct Answer:} \colorbox{green!30}{259.808 meter} 
\end{tcolorbox}

\subsection{Language Coverage in the {{\dataname}} Benchmark}
\label{appendix:language_stat}
This appendix provides detailed statistics of the languages included in the {{\dataname}} benchmark.
\begin{table}[!ht]
  \centering
  \begin{adjustbox}{width=0.70\textwidth}
    \begin{tabular}{lllc c}
      \toprule
      \rowcolor{gray!15} \textbf{Code} & \textbf{Language} & \textbf{Language Family} & \textbf{Script} & \textbf{Speakers (M)} \\
      \midrule
      en & English      & Indo-European & Latin         & $\approx$ 1{,}500 \\
      ar & Arabic       & Afro-Asiatic  & Arabic (RTL)  & $\approx$ 420 \\
      bn & Bangla       & Indo-European & Bengali       & $\approx$ 300 \\
      fr & French       & Indo-European & Latin         & $\approx$ 300 \\
      sw & Swahili      & Niger-Congo   & Latin         & $\approx$ 200 \\
      fa & Persian      & Indo-European & Arabic (RTL)  & $\approx$ 130 \\
      tr & Turkish      & Turkic        & Latin         & $\approx$ 85 \\
      ha & Hausa        & Afro-Asiatic  & Latin         & $\approx$ 80 \\
      gu & Gujarati     & Indo-European & Gujarati      & $\approx$ 60 \\
      am & Amharic      & Afro-Asiatic  & Ethiopic      & $\approx$ 57 \\
      kk & Kazakh       & Turkic        & Cyrillic      & $\approx$ 15 \\
      fi & Finnish      & Uralic        & Latin         & $\approx$ 6 \\
      lt & Lithuanian   & Indo-European & Latin         & $\approx$ 3 \\
      \midrule
      \multicolumn{4}{l}{\textbf{Total Speaker Reach}} & $\approx$ 3.15 billion \\
      \bottomrule
    \end{tabular}
  \end{adjustbox}
  \caption{Detailed Statistics of the Languages in {{\dataname}} Benchmark}
  \label{tab:mathmist_languages}
\end{table}
Table \ref{tab:mathmist_languages} summarizes the linguistic diversity of the dataset, reporting each language’s ISO code, language family, writing script, and an approximate estimate of the number of native and second-language speakers worldwide. The selected languages encompass various language families and writing systems, including Indo-European, Afro-Asiatic, Niger-Congo, Turkic, and Uralic. They represent both left-to-right and right-to-left scripts, as well as Semitic and Bantu morphologies. Together, these languages represent an estimated total speaker reach of approximately 3.15 billion, highlighting the broad global coverage of {{\dataname}}. This diversity enables a systematic evaluation of multilingual mathematical reasoning across typologically and script-wise heterogeneous languages.

\subsection{LLM-as-a-Judge: Evaluation Protocol and Scoring Rubric}
\label{appendix:scoring_rubric}
We conducted a human review with an SME (subject-matter expert), which is standard for mathematical evaluation requiring domain precision. Table \ref{tab:scoring_rubric} shows a 0 to 100 human scoring rubric that covers correctness, alignment of reasoning, and clarity.
The SME evaluated and ensured uniformity without inter-annotator variability. Additionally, the SME performed back-translation checks using Google Translate and DeepL to verify semantic fidelity across languages. The LLM-as-a-Judge achieved a Human Consistency Score ranging from 91.39/100 to 94.8/100 across non-MCQ experiments. This approach aligns closely with expert reasoning. Minor deductions were mostly due to explanations that were correct but slightly under-specified. This analysis reinforces the reliability of the judge’s decisions and their alignment with expert human understanding.

\subsection{Analysis of Scaling Effects on the Qwen Model Family}
\label{appendix:scaling_effect}
We conducted an ablation study with the Qwen 3 model family, ranging from 0.6B to 14B parameters, to analyze scaling effects on mathematical reasoning in English and Bangla. Our results confirm a strong positive correlation between parameter size and performance, particularly in English, where accuracy improved from 26.85\% at 0.6B to 45.95\% at 14B. Larger models significantly reduce the performance gap between English and Bangla. The smallest model (0.6 billion parameters) showed a 7\% advantage for English over Bangla, while the largest model (14 billion parameters) resulted in nearly equal performance, with only a 1\% difference. Furthermore, the ability to solve logical proofs improved with model size. In Bangla, success rates nearly doubled from 38\% with the smaller model to 70\% with the larger model. We have included a detailed breakdown of these results in the table \ref{tab:qwen_scaling_ablation} below.
\begin{table*}[!ht]
\centering
\begin{adjustbox}{width=\textwidth}
\begin{tabular}{lcccccccc}
\toprule
\multirow{2}{*}{\textbf{Model}} &
\multicolumn{4}{c}{\textbf{Bangla Accuracy (\%)}} &
\multicolumn{4}{c}{\textbf{English Accuracy (\%)}} \\
\cmidrule(lr){2-5} \cmidrule(lr){6-9}
& Numerical & Symbolic & Proof & Overall
& Numerical & Symbolic & Proof & Overall \\
\midrule
\verb|Qwen 0.6B|
& 12.93 & 18.97 & 38.78 & 19.72
& 16.61 & 21.72 & 59.29 & 26.85 \\

\verb|Qwen 1.7B|
& 12.34 {\color{red}{($\downarrow$0.59)}}
& 23.10 {\color{darkgreen}{($\uparrow$4.13)}}
& 52.88 {\color{darkgreen}{($\uparrow$14.10)}}
& 23.25 {\color{darkgreen}{($\uparrow$3.53)}}
& 26.69 {\color{darkgreen}{($\uparrow$10.08)}}
& 21.03 {\color{red}{($\downarrow$0.69)}}
& 54.17 {\color{red}{($\downarrow$5.12)}}
& 31.49 {\color{darkgreen}{($\uparrow$4.64)}} \\

\verb|Qwen 4B|
& \colorbox{green!20}{\textbf{64.53}} {\color{darkgreen}{($\uparrow$51.60)}}
& \colorbox{green!20}{\textbf{61.72}} {\color{darkgreen}{($\uparrow$42.75)}}
& 56.41 {\color{darkgreen}{($\uparrow$17.63)}}
& \colorbox{green!20}{\textbf{62.21}} {\color{darkgreen}{($\uparrow$42.49)}}
& 32.62 {\color{darkgreen}{($\uparrow$16.01)}}
& 26.90 {\color{darkgreen}{($\uparrow$5.18)}}
& 60.26 {\color{darkgreen}{($\uparrow$0.97)}}
& 37.44 {\color{darkgreen}{($\uparrow$10.59)}} \\

\verb|Qwen 8B|
& 54.92 {\color{darkgreen}{($\uparrow$41.99)}}
& 47.24 {\color{darkgreen}{($\uparrow$28.27)}}
& 64.10 {\color{darkgreen}{($\uparrow$25.32)}}
& 55.36 {\color{darkgreen}{($\uparrow$35.64)}}
& \colorbox{green!20}{\textbf{37.49}} {\color{darkgreen}{($\uparrow$20.88)}}
& 42.76 {\color{darkgreen}{($\uparrow$21.04)}}
& 61.22 {\color{darkgreen}{($\uparrow$1.93)}}
& 43.67 {\color{darkgreen}{($\uparrow$16.82)}} \\

\verb|Qwen 14B|
& 36.30 {\color{darkgreen}{($\uparrow$23.37)}}
& 42.76 {\color{darkgreen}{($\uparrow$23.79)}}
& \colorbox{green!20}{\textbf{70.19}} {\color{darkgreen}{($\uparrow$31.41)}}
& 44.91 {\color{darkgreen}{($\uparrow$25.19)}}
& 34.40 {\color{darkgreen}{($\uparrow$17.79)}}
& \colorbox{green!20}{\textbf{47.59}} {\color{darkgreen}{($\uparrow$25.87)}}
& \colorbox{green!20}{\textbf{75.64}} {\color{darkgreen}{($\uparrow$16.35)}}
& \colorbox{green!20}{\textbf{45.95}} {\color{darkgreen}{($\uparrow$19.10)}} \\
\bottomrule
\end{tabular}
\end{adjustbox}

\caption{
Scaling effects of the \texttt{Qwen-3} model family (0.6B–14B) on mathematical reasoning in Bangla and English. Relative changes (\textcolor{darkgreen}{$\uparrow$} / \textcolor{red}{$\downarrow$}) indicate performance differences with respect to the smallest model (\texttt{Qwen-3:0.6B}) within the same language and category. \colorbox{green!20}{Green} highlights the highest accuracy achieved in each column. Results demonstrate a strong positive correlation between model size and reasoning performance, particularly for proof-based tasks and overall accuracy.
}
\label{tab:qwen_scaling_ablation}
\end{table*}
We marked unexpectedly high numerical reasoning spikes in the \texttt{Qwen-3:4B} Bangla checkpoint, presumably due to specific data mixture variance in that release. But the general trend confirms that larger models offer stable cross-lingual alignment.

\subsection{Breakdown of Quantitative Analysis}
\label{appendix:qualitative_analysis}
The tables detail an evaluation of LLM mathematical reasoning across English and Bangla, focusing on the efficacy of Chain-of-Thought (CoT) prompting and the problem of linguistic interference.
The data reveals that CoT generally boosts performance (Table \ref{tab:zero_cot_en_bn_category_wise}), especially for smaller models, but high-performing models show minimal or negative gains in Bangla, indicating linguistic brittleness. 
\begin{table*}[t]
  \centering
  \begin{adjustbox}{width=\textwidth}
    \begin{tabular}{lcccccllll}
      \toprule
      \multirow{3}{*}{\textbf{Models}} & \multirow{3}{*}{\textbf{\#Param}} 
      & \multicolumn{8}{c}{\textbf{Bangla}} \\
      \cmidrule(lr){3-10}
       & & \multicolumn{4}{c|}{\textbf{Zero-Shot}} & \multicolumn{4}{c}{\textbf{CoT}} \\
      \cmidrule(lr){3-6} \cmidrule(lr){7-10}
       & & Numerical & Symbolic & Proof & Overall & Numerical & Symbolic & Proof & Overall \\
      \midrule
      \verb|Qwen| & 8B 
      & 54.92 & 47.24 & 64.10 & 55.36
      & 73.67 {\color{darkgreen}{($\uparrow$18.75)}} & 68.28 {\color{darkgreen}{($\uparrow$21.04)}} & \colorbox{green!20}{\textbf{71.47}} {\color{darkgreen}{($\uparrow$7.37)}} & 72.11 {\color{darkgreen}{($\uparrow$16.75)}} \\

      \verb|GPT-OSS| & 20B
      &  79.00 & \colorbox{blue!15}{\textbf{73.45}} & 62.50 & 74.33
      & \colorbox{red!20}{\textbf{78.29}} {\color{red}{($\downarrow$0.71)}} & \colorbox{blue!15}{\textbf{71.72}} {\color{red}{($\downarrow$1.73)}} & 66.03 {\color{darkgreen}{($\uparrow$3.53)}} & \colorbox{magenta!15}{\textbf{75.22}} {\color{darkgreen}{($\uparrow$0.89)}} \\

      \verb|DeepSeek R1| & 7B
      & 32.38 & 35.17 & \colorbox{green!20}{\textbf{67.31}} & 40.48
      & 61.09 {\color{darkgreen}{($\uparrow$28.71)}} & 54.83 {\color{darkgreen}{($\uparrow$19.66)}} & 66.03 {\color{red}{($\downarrow$1.28)}} & 60.90 {\color{darkgreen}{($\uparrow$20.42)}} \\

      \verb|Mathstral| & 7B
      & 24.08 & 25.52 & 56.73 & 31.42
      & 35.47 {\color{darkgreen}{($\uparrow$11.39)}} & 37.59 {\color{darkgreen}{($\uparrow$12.07)}} & 59.94 {\color{darkgreen}{($\uparrow$3.21)}} & 41.18 {\color{darkgreen}{($\uparrow$9.76)}} \\

      \verb|Gemini-2.5-Flash-Lite| & N/A
      & \colorbox{red!20}{\textbf{79.95}} & 71.72 & 65.06 & \colorbox{magenta!15}{\textbf{75.09}}
      & 73.07 {\color{red}{($\downarrow$6.88)}} & 70.69 {\color{red}{($\downarrow$1.03)}} & 65.38 {\color{red}{($\downarrow$0.68)}} & 71.38 {\color{red}{($\downarrow$3.71)}} \\
      \midrule[\heavyrulewidth]
      
      \multirow{3}{*}{\textbf{Models}} & \multirow{3}{*}{\textbf{\#Param}} 
      & \multicolumn{8}{c}{\textbf{English}} \\
      \cmidrule(lr){3-10}
       & & \multicolumn{4}{c|}{\textbf{Zero-Shot}} & \multicolumn{4}{c}{\textbf{CoT}} \\
      \cmidrule(lr){3-6} \cmidrule(lr){7-10}
       & & Numerical & Symbolic & Proof & Overall & Numerical & Symbolic & Proof & Overall \\
      \midrule
      \verb|Qwen| & 8B
      & 37.49 & 42.76 & 61.22 & 43.67
      & 78.05 {\color{darkgreen}{($\uparrow$40.56)}} & 72.76 {\color{darkgreen}{($\uparrow$30.00)}} & \colorbox{green!20}{\textbf{72.44}} {\color{darkgreen}{($\uparrow$11.22)}} & 75.78 {\color{darkgreen}{($\uparrow$32.11)}} \\

      \verb|GPT-OSS| & 20B
      & \colorbox{red!20}{\textbf{79.12}} & \colorbox{blue!15}{\textbf{75.52}} & \colorbox{green!20}{\textbf{68.59}} & \colorbox{magenta!15}{\textbf{76.12}}
      & \colorbox{red!20}{\textbf{80.90}} {\color{darkgreen}{($\uparrow$1.78)}} & \colorbox{blue!15}{\textbf{75.52}} {\color{darkgreen}{($\uparrow$0.00)}} & 70.51 {\color{darkgreen}{($\uparrow$1.92)}} & \colorbox{magenta!15}{\textbf{77.58}} {\color{darkgreen}{($\uparrow$1.46)}} \\

      \verb|DeepSeek R1| & 7B
      & 41.28 & 40.34 & 65.06 & 46.23
      & 66.79 {\color{darkgreen}{($\uparrow$25.51)}} & 60.34 {\color{darkgreen}{($\uparrow$19.99)}} & 63.78 {\color{red}{($\downarrow$1.28)}} & 64.84 {\color{darkgreen}{($\uparrow$18.61)}} \\

      \verb|Mathstral| & 7B
      & 21.23 & 27.93 & 53.85 & 29.62
      & 40.33 {\color{darkgreen}{($\uparrow$19.10)}} & 38.97 {\color{darkgreen}{($\uparrow$11.04)}} & 64.42 {\color{darkgreen}{($\uparrow$10.57)}} & 45.26 {\color{darkgreen}{($\uparrow$15.64)}} \\

      \verb|Gemini-2.5-Flash-Lite| & N/A
      & 64.18 & 59.31 & 59.29 & 62.15
      & 74.14 {\color{darkgreen}{($\uparrow$9.96)}} & 71.38 {\color{darkgreen}{($\uparrow$12.07)}} & 65.06 {\color{darkgreen}{($\uparrow$5.77)}} & 71.63 {\color{darkgreen}{($\uparrow$9.48)}} \\

      \bottomrule
    \end{tabular}
  \end{adjustbox}
  \caption{Performance breakdown (\texttt{Numerical}, \texttt{Symbolic}, \texttt{Proof}, \texttt{Overall}) for Zero-Shot versus CoT on the \texttt{\dataname} BN-EN parallel corpus. The maximum values in each language and setting are highlighted as follows: \colorbox{red!20}{\texttt{Numerical}}, \colorbox{blue!15}{\texttt{Symbolic}}, \colorbox{green!20}{\texttt{Proof}}, and \colorbox{magenta!15}{\texttt{Overall}}. Upward arrows, indicated by \textcolor{darkgreen}{$\uparrow$}, represent improvements, while downward arrows, shown as \textcolor{red}{$\downarrow$}, indicate a decrease in performance. Arrows in parentheses reflect changes relative to the Zero-Shot values.}
  \label{tab:zero_cot_en_bn_category_wise}
\end{table*}
This pervasive linguistic inconsistency (Table \ref{tab:code_switch_breakdown}) identifies a major challenge to cross-lingual robustness and instruction fidelity. Table \ref{tab:translation_acc} presents results for all evaluated models, categorized by language and problem types. Overall, GPT-OSS-20B stands out as the top performer, achieving the highest scores in both numerical and symbolic categories across most languages. It also exhibits competitive proof accuracy, leading in languages such as English, French, Kazakh, Finnish, Lithuanian, Turkish, Persian, and many low-resource languages. Gemini-2.5-Flash-Lite ranks as a solid second-tier model, achieving the next-best overall and proof accuracy averages. In some cases, it even surpasses GPT-OSS; for example, it demonstrates superior numerical performance in Bangla and better proof scores in Arabic, Gujarati, and Amharic. Qwen-8B and DeepSeek R1 7B are positioned in the middle tier. Both models show relatively strong proof capabilities in several languages. 
\begin{table*}[!ht]
  \centering
  \begin{adjustbox}{width=\textwidth}
    \begin{tabular}{lcccccllll}
      \toprule
      \multirow{3}{*}{\textbf{Models}} & \multirow{3}{*}{\textbf{\#Param}} 
      & \multicolumn{8}{c}{\textbf{Question Bangla}} \\
      \cmidrule(lr){3-10}
       & & \multicolumn{4}{c|}{\textbf{CoT English}} & \multicolumn{4}{c}{\textbf{CoT Bangla}} \\
      \cmidrule(lr){3-6} \cmidrule(lr){7-10}
       & & Numerical & Symbolic & Proof & Overall & Numerical & Symbolic & Proof & Overall \\
      \midrule
      \verb|GPT-OSS| & 20B
      & \colorbox{red!20}{\textbf{79.48}} & \colorbox{blue!15}{\textbf{70.00}} & 67.95 & \colorbox{magenta!15}{\textbf{75.09}}
      & 76.51 {\color{red}{($\downarrow$2.97)}} & 71.03 {\color{darkgreen}{($\uparrow$1.03)}} & 66.67 {\color{red}{($\downarrow$1.28)}} & 73.29 {\color{red}{($\downarrow$1.80)}} \\

      \verb|Qwen| & 8B 
      & 73.67 & 68.28 & \colorbox{green!20}{\textbf{71.47}} & 72.11
      & 72.24 {\color{red}{($\downarrow$1.43)}} & 70.69 {\color{darkgreen}{($\uparrow$2.41)}} & 70.19 {\color{red}{($\downarrow$1.28)}} & 71.49 {\color{red}{($\downarrow$0.62)}} \\

      \verb|DeepSeek R1| & 7B
      & 61.09 & 54.83 & 66.03 & 60.90
      & 52.43 {\color{red}{($\downarrow$8.66)}} & 53.45 {\color{red}{($\downarrow$1.38)}} & 63.78 {\color{red}{($\downarrow$2.25)}} & 55.09 {\color{red}{($\downarrow$5.81)}} \\

      \verb|Mathstral| & 7B
      & 35.47 & 37.59 & 59.94 & 41.18
      & 30.01 {\color{red}{($\downarrow$5.46)}} & 30.34 {\color{red}{($\downarrow$7.25)}} & 51.60 {\color{red}{($\downarrow$8.34)}} & 34.74 {\color{red}{($\downarrow$6.44)}} \\

      \midrule[\heavyrulewidth]
      
      \multirow{3}{*}{\textbf{Models}} & \multirow{3}{*}{\textbf{\#Param}} 
      & \multicolumn{8}{c}{\textbf{Question English}} \\
      \cmidrule(lr){3-10}
       & & \multicolumn{4}{c|}{\textbf{CoT English}} & \multicolumn{4}{c}{\textbf{CoT Bangla}} \\
      \cmidrule(lr){3-6} \cmidrule(lr){7-10}
       & & Numerical & Symbolic & Proof & Overall & Numerical & Symbolic & Proof & Overall \\
      \midrule
      \verb|GPT-OSS| & 20B
      & \colorbox{red!20}{\textbf{79.36}} & \colorbox{blue!15}{\textbf{76.55}} & \colorbox{green!20}{\textbf{73.72}} & \colorbox{magenta!15}{\textbf{77.58}}
      & 79.12 {\color{red}{($\downarrow$0.24)}} & 75.52 {\color{red}{($\downarrow$1.03)}} & 64.42 {\color{red}{($\downarrow$9.30)}} & 75.22 {\color{red}{($\downarrow$2.36)}} \\

      \verb|Qwen| & 8B
      & 78.05 & 72.76 & 72.44 & 75.78
      & 76.75 {\color{red}{($\downarrow$1.30)}} & 73.10 {\color{darkgreen}{($\uparrow$0.34)}} & 66.99 {\color{red}{($\downarrow$5.45)}} & 73.91 {\color{red}{($\downarrow$1.87)}} \\

      \verb|DeepSeek R1| & 7B
      & 66.79 & 60.34 & 63.78 & 64.84
      & 58.60 {\color{red}{($\downarrow$8.19)}} & 59.31 {\color{red}{($\downarrow$1.03)}} & 62.50 {\color{red}{($\downarrow$1.28)}} & 59.58 {\color{red}{($\downarrow$5.26)}} \\

      \verb|Mathstral| & 7B
      & 40.33 & 38.97 & 64.42 & 45.26
      & 32.86 {\color{red}{($\downarrow$7.47)}} & 28.62 {\color{red}{($\downarrow$10.35)}} & 55.13 {\color{red}{($\downarrow$9.29)}} & 36.82 {\color{red}{($\downarrow$8.44)}} \\

      \bottomrule
    \end{tabular}
  \end{adjustbox}
  \caption{Accuracy breakdown (\%) across mathematical problem types underneath English–Bangla code-switching reasoning. Each block contrasts aligned and cross-lingual settings across LLMs of different scales. Maximum scores are highlighted by category: \colorbox{red!20}{\texttt{Numerical}}, \colorbox{blue!15}{\texttt{Symbolic}}, \colorbox{green!20}{\texttt{Proof}}, and \colorbox{magenta!15}{\texttt{Overall}}. Arrows (\textcolor{darkgreen}{$\uparrow$}/\textcolor{red}{$\downarrow$}) indicate performance change relative to the chain-of-thought (CoT) in English.}
  \label{tab:code_switch_breakdown}
\end{table*}
Notably, Qwen performs well in various European languages, while DeepSeek achieves the highest proof score for Bangla. However, both fall short compared to the top two models in terms of numerical and symbolic tasks. Mathstral 7B is the weakest model overall, particularly lacking in numerical and overall accuracy, with consistently lower scores across both high- and low-resource languages. In summary, GPT-OSS-20B excels in symbolic and numerical categories across languages. Gemini serves as a robust alternative with specific strengths in proof accuracy for certain languages. Qwen and DeepSeek are mid-tier models with selective strengths in proof, while Mathstral underperforms compared to the other models. Moreover, the language composition tables (\ref{tab:gptoss20b_bn_bn_lang_stats}) expose widespread code-switching reasoning during Bangla CoT: smaller models (DeepSeek R1, Mathstral) exhibit  "Mixed" language outputs, injecting English, Cyrillic, etc. (Table \ref{tab:deepseek_bn_bn_lang_stats}).


\begin{table*}[t]
\tiny
\centering
\setlength{\tabcolsep}{3.5pt}
\begin{tabular}{l l c c c c c c c c c c c c c }
\toprule
\rowcolor{gray!15} \textbf{Model} & \textbf{Metric} & \textbf{en} & \textbf{bn} & \textbf{fr} & \textbf{kk} & \textbf{fi} & \textbf{lt} & \textbf{tr} & \textbf{fa} & \textbf{ar} & \textbf{sw} & \textbf{ha} & \textbf{gu} & \textbf{am} \\
\midrule
\rowcolor{blue!10} \cellcolor{white} & \textbf{Overall} & 43.67\% & 55.36\% & 52.11\% & 47.4\% & 50.73\% & 45.4\% & 49.62\% & 68.1\% & 32.73\% & 24.98\% & 23.39\% & 40.76\% & 23.39\% \\
\rowcolor{red!5} \cellcolor{white} & \textbf{Numerical} & 37.49\% & 54.92\% & 47.45\% & 43.18\% & 45.2\% & 38.2\% & 44.6\% & 67.14\% & 24.56\% & 18.86\% & 18.27\% & 32.15\% & 19.69\% \\
\rowcolor{green!5} \cellcolor{white} & \textbf{Proof} & 61.22\% & 64.1\% & 64.1\% & 58.65\% & 69.23\% & 65.71\% & 61.22\% & 66.35\% & 50.96\% & 37.18\% & 36.54\% & 62.5\% & 30.45\% \\
\rowcolor{orange!10} \cellcolor{white} \multirow{-4}{*}{\textbf{Qwen-8B}} & \textbf{Symbolic} & 42.76\% & 47.24\% & 52.76\% & 47.59\% & 46.9\% & 44.48\% & 51.72\% & 72.76\% & 36.9\% & 29.66\% & 24.14\% & 42.41\% & 26.55\% \\
\addlinespace
\rowcolor{blue!10} \cellcolor{white} & \textbf{Overall} & 76.12\% & 74.33\% & 75.78\% & 73.77\% & 73.56\% & 73.08\% & 74.05\% & 72.04\% & 73.08\% & 65.67\% & 67.89\% & 72.8\% & 63.11\% \\
\rowcolor{red!5} \cellcolor{white} & \textbf{Numerical} & 79.12\% & 79.0\% & 75.92\% & 73.07\% & 72.95\% & 70.46\% & 71.77\% & 71.53\% & 74.26\% & 65.72\% & 67.62\% & 75.21\% & 64.06\% \\
\rowcolor{green!5} \cellcolor{white} & \textbf{Proof} & 68.59\% & 62.5\% & 75.0\% & 76.28\% & 74.04\% & 77.56\% & 77.56\% & 71.15\% & 66.67\% & 66.35\% & 68.27\% & 65.38\% & 58.65\% \\
\rowcolor{orange!10} \cellcolor{white} \multirow{-4}{*}{\textbf{GPT-OSS-20B}} & \textbf{Symbolic} & 75.52\% & 73.45\% & 76.21\% & 73.1\% & 74.83\% & 75.86\% & 76.9\% & 74.48\% & 76.55\% & 64.83\% & 68.28\% & 73.79\% & 65.17\% \\
\addlinespace
\rowcolor{blue!10} \cellcolor{white} & \textbf{Overall} & 46.22\% & 40.48\% & 47.06\% & 33.43\% & 38.06\% & 37.16\% & 42.01\% & 51.42\% & 40.83\% & 29.34\% & 25.67\% & 35.5\% & 28.17\% \\
\rowcolor{red!5} \cellcolor{white} & \textbf{Numerical} & 41.28\% & 32.38\% & 42.11\% & 29.18\% & 29.3\% & 30.01\% & 35.23\% & 46.98\% & 36.65\% & 23.72\% & 19.22\% & 25.74\% & 21.95\% \\
\rowcolor{green!5} \cellcolor{white} & \textbf{Proof} & 65.06\% & 67.31\% & 65.38\% & 42.31\% & 59.62\% & 58.01\% & 61.54\% & 62.5\% & 52.88\% & 43.27\% & 41.03\% & 59.29\% & 41.99\% \\
\rowcolor{orange!10} \cellcolor{white} \multirow{-4}{*}{\textbf{DeepSeek R1 7B}} & \textbf{Symbolic} & 40.34\% & 35.17\% & 41.72\% & 36.21\% & 40.34\% & 35.52\% & 40.69\% & 52.41\% & 40.0\% & 30.69\% & 27.93\% & 38.28\% & 31.38\% \\
\addlinespace
\rowcolor{blue!10} \cellcolor{white} & \textbf{Overall} & 29.63\% & 31.43\% & 31.28\% & 19.24\% & 17.92\% & 22.91\% & 23.04\% & 25.54\% & 22.77\% & 15.09\% & 8.86\% & 21.66\% & 11.21\% \\
\rowcolor{red!5} \cellcolor{white} & \textbf{Numerical} & 21.23\% & 24.08\% & 21.71\% & 11.27\% & 9.96\% & 14.47\% & 13.05\% & 16.61\% & 14.95\% & 10.32\% & 5.1\% & 13.4\% & 7.71\% \\
\rowcolor{green!5} \cellcolor{white} & \textbf{Proof} & 53.85\% & 56.73\% & 58.33\% & 33.65\% & 39.42\% & 45.51\% & 49.04\% & 45.51\% & 40.38\% & 26.92\% & 16.03\% & 44.87\% & 14.1\% \\
\rowcolor{orange!10} \cellcolor{white} \multirow{-4}{*}{\textbf{Mathstral 7B}} & \textbf{Symbolic} & 27.93\% & 25.52\% & 30.0\% & 26.9\% & 17.93\% & 23.1\% & 24.14\% & 30.0\% & 26.55\% & 16.21\% & 12.07\% & 20.69\% & 18.28\% \\
\addlinespace
\rowcolor{blue!10} \cellcolor{white} & \textbf{Overall} & 62.15\% & 75.12\% & 63.53\% & 53.15\% & 56.19\% & 52.25\% & 53.36\% & 50.59\% & 50.8\% & 46.92\% & 40.48\% & 48.3\% & 44.78\% \\
\rowcolor{red!5} \cellcolor{white} & \textbf{Numerical} & 64.18\% & 79.95\% & 60.62\% & 48.04\% & 53.26\% & 47.33\% & 45.55\% & 43.3\% & 43.89\% & 44.01\% & 34.16\% & 39.5\% & 35.71\% \\
\rowcolor{green!5} \cellcolor{white} & \textbf{Proof} & 59.29\% & 65.06\% & 68.27\% & 61.54\% & 61.22\% & 66.99\% & 69.87\% & 68.27\% & 68.91\% & 55.45\% & 54.17\% & 69.55\% & 65.38\% \\
\rowcolor{orange!10} \cellcolor{white} \multirow{-4}{*}{\textbf{Gemini-2.5-Flash-Lite}} & \textbf{Symbolic} & 59.31\% & 71.72\% & 66.9\% & 58.97\% & 59.31\% & 50.69\% & 58.28\% & 52.76\% & 51.38\% & 46.21\% & 44.14\% & 51.03\% & 48.97\% \\
\addlinespace
\bottomrule
\end{tabular}
\caption{Full Result by Category of {\dataname}. Here, en: English, bn: Bangla, fr: French, kk: Kazakh, fi: Finnish, lt: Lithuanian, tr: Turkish, fa: Persian, ar: Arabic, sw: Swahili, ha: Hausa, gu: Gujarati, am: Amharic.}
\label{tab:translation_acc}
\end{table*}

\begin{table*}[!ht]
  \centering
  \begin{adjustbox}{width=0.80\textwidth}
    \begin{tabular}{lccccc}
      \toprule
      \multicolumn{6}{c}{\textbf{Language presence in CoT solutions — \texttt{GPT-OSS-20B} (Question: Bangla, CoT: Bangla)}} \\
      \midrule
      \multicolumn{3}{c}{\textbf{Aggregate language counts}} & \multicolumn{3}{c}{\textbf{Language combinations in CoT}} \\
      \cmidrule(lr){1-3} \cmidrule(lr){4-6}
      \textbf{Language} & \textbf{Count} & \textbf{Proportion (\%)} & \textbf{Combination} & \textbf{Count} & \textbf{Proportion (\%)} \\
      \midrule
      Bangla   & 1188 & 82.2 & Bangla + English              & 225 & 94.9 \\
      English  &   21 &  1.5 & Bangla + Chinese              &   8 &  3.4 \\
      Chinese  &    0 &  0.0 & Bangla + Cyrillic + English   &   1 &  0.4 \\
      Mixed    &  236 & 16.3 & Bangla + Chinese + English    &   1 &  0.4 \\
      \multicolumn{3}{c}{}     & Bangla + English + Greek      &   1 &  0.4 \\
      \bottomrule
    \end{tabular}
  \end{adjustbox}
  \caption{Language composition in chain-of-thought (CoT) outputs generated by the \texttt{GPT-OSS-20B} model when prompted with Bangla questions and instructed to reason in Bangla. The table summarizes aggregate counts and proportions of detected languages, alongside mixed-language combinations observed in CoT reasoning traces. Notably, \texttt{Mixed} CoTs remain relatively limited, reflecting the model’s strong linguistic alignment and controlled cross-lingual consistency.}
  \label{tab:gptoss20b_bn_bn_lang_stats}
\end{table*}

\begin{table*}[!ht]
  \centering
  \begin{adjustbox}{width=0.80\textwidth}
    \begin{tabular}{lccccc}
      \toprule
      \multicolumn{6}{c}{\textbf{Language presence in CoT solutions — \texttt{GPT-OSS-20B} (Question: English, CoT: Bangla)}} \\
      \midrule
      \multicolumn{3}{c}{\textbf{Aggregate language counts}} & \multicolumn{3}{c}{\textbf{Language combinations in CoT}} \\
      \cmidrule(lr){1-3} \cmidrule(lr){4-6}
      \textbf{Language} & \textbf{Count} & \textbf{Proportion (\%)} & \textbf{Combination} & \textbf{Count} & \textbf{Proportion (\%)} \\
      \midrule
      Bangla   & 1157 & 80.1 & Bangla + English             & 267 & 99.3 \\
      English  &   19 &  1.3 & Bangla + Chinese + English   &   1 &  0.4 \\
      Chinese  &    0 &  0.0 & Bangla + Chinese             &   1 &  0.4 \\
      Mixed    &  269 & 18.6 & ---                          & --- & --- \\
      \bottomrule
    \end{tabular}
  \end{adjustbox}
  \caption{Aggregate language distribution in chain-of-thought (CoT) outputs for the \texttt{GPT-OSS-20B} model when prompted with English questions and instructed to reason in Bangla. The table summarizes the overall occurrence of individual languages and mixed-language combinations observed in CoT solutions. Here, \texttt{Mixed} indicates CoTs that contain linguistic tokens from multiple languages, revealing the extent of code-switching during multilingual reasoning.}
  \label{tab:gptoss_en_bn_lang_stats}
\end{table*}

\begin{table*}[!ht]
  \centering
  \begin{adjustbox}{width=.90\textwidth}
    \begin{tabular}{lccccc}
      \toprule
      \multicolumn{6}{c}{\textbf{Language presence in CoT solutions — \texttt{DeepSeek R1-7B} (Question: Bangla, CoT: Bangla)}} \\
      \midrule
      \multicolumn{3}{c}{\textbf{Aggregate language counts}} & \multicolumn{3}{c}{\textbf{Language combinations in CoT (percent of Mixed)}} \\
      \cmidrule(lr){1-3} \cmidrule(lr){4-6}
      \textbf{Language} & \textbf{Count} & \textbf{Proportion (\%)} & \textbf{Combination} & \textbf{Count} & \textbf{Proportion (\%) } \\
      \midrule
      Bangla   & 523 & 36.2 & Bangla + English                          & 689 & 83.1 \\
      English  &  93 &  6.4 & Bangla + Cyrillic + English               &  42 &  5.1 \\
      Chinese  &   0 &  0.0 & Bangla + Chinese + English                &  39 &  4.7 \\
      Mixed    & 829 & 57.4 & Bangla + Chinese                          &  28 &  3.4 \\
               &     &      & Bangla + English + Greek                  &  13 &  1.6 \\
               &     &      & Arabic + Bangla + English                 &   8 &  1.0 \\
               &     &      & Bangla + Chinese + Cyrillic + English     &   3 &  0.4 \\
               &     &      & Bangla + Chinese + Cyrillic               &   2 &  0.2 \\
               &     &      & Arabic + Bangla + Chinese + English + Japanese & 1 & 0.1 \\
               &     &      & Arabic + Bangla + Chinese + English       &   1 &  0.1 \\
               &     &      & Arabic + Bangla + Cyrillic + English      &   1 &  0.1 \\
               &     &      & Arabic + Bangla + Chinese                 &   1 &  0.1 \\
               &     &      & Arabic + Bangla + English + Greek         &   1 &  0.1 \\
      \bottomrule
    \end{tabular}
  \end{adjustbox}
  \caption{Aggregate distribution of languages in chain-of-thought (CoT) outputs for the \texttt{DeepSeek R1-7B} model when both the question and CoT were instructed in Bangla. The table summarizes the overall frequency and proportion of individual languages, as well as the composition and percentage of mixed-language CoTs. The majority of mixed outputs involve Bangla–English combinations, with occasional incorporation of additional scripts such as Cyrillic, Chinese, Arabic, Greek, and Japanese, reflecting the model’s multilingual interference during reasoning.}
  \label{tab:deepseek_bn_bn_lang_stats}
\end{table*}

\begin{table*}[!ht]
  \centering
  \begin{adjustbox}{width=.90\textwidth}
    \begin{tabular}{lccccc}
      \toprule
      \multicolumn{6}{c}{\textbf{Language presence in CoT solutions — \texttt{DeepSeek-R1-7B} (Question: English, CoT: Bangla)}} \\
      \midrule
      \multicolumn{3}{c}{\textbf{Aggregate language counts}} & \multicolumn{3}{c}{\textbf{Language combinations in CoT}} \\
      \cmidrule(lr){1-3} \cmidrule(lr){4-6}
      \textbf{Language} & \textbf{Count} & \textbf{Proportion (\%)} & \textbf{Combination} & \textbf{Count} & \textbf{Proportion (\%)} \\
      \midrule
      Bangla   & 542 & 37.5 & Bangla + English                        & 611 & 78.7 \\
      English  & 126 &  8.7 & Bangla + Chinese + English              &  62 &  8.0 \\
      Chinese  &   1 &  0.1 & Bangla + Cyrillic + English             &  49 &  6.3 \\
      Mixed    & 776 & 53.7 & Bangla + Chinese                        &  10 &  1.3 \\
               &     &      & Arabic + Bangla + English               &   8 &  1.0 \\
               &     &      & Bangla + English + Greek                &   8 &  1.0 \\
               &     &      & Bangla + Cyrillic                       &   4 &  0.5 \\
               &     &      & Bangla + Chinese + Cyrillic             &   4 &  0.5 \\
               &     &      & Bangla + English + Korean               &   3 &  0.4 \\
               &     &      & Bangla + Chinese + Cyrillic + English   &   3 &  0.4 \\
               &     &      & Bangla + English + Japanese             &   2 &  0.3 \\
               &     &      & Bangla + English + Thai                 &   2 &  0.3 \\
               &     &      & Bangla + Cyrillic + English + Greek     &   2 &  0.3 \\
               &     &      & Bangla + Chinese + English + Korean     &   1 &  0.1 \\
               &     &      & Arabic + Bangla + Chinese + English     &   1 &  0.1 \\
               &     &      & Bangla + Korean                         &   1 &  0.1 \\
               &     &      & Arabic + Bangla + Cyrillic + English    &   1 &  0.1 \\
               &     &      & Arabic + Bangla                         &   1 &  0.1 \\
               &     &      & Bangla + English + Greek + Thai         &   1 &  0.1 \\
               &     &      & Arabic + Bangla + English + Greek       &   1 &  0.1 \\
               &     &      & Bangla + Chinese + English + Japanese   &   1 &  0.1 \\
      \bottomrule
    \end{tabular}
  \end{adjustbox}
  \caption{Aggregate language distribution in chain-of-thought (CoT) outputs for the \texttt{DeepSeek-R1-7B} model when prompted with English questions and instructed to reason in Bangla. The table reports overall counts and proportions of languages present in the CoT, along with detailed statistics of mixed-language combinations. \texttt{Mixed} indicates CoTs containing multiple linguistic scripts or tokens from different languages, reflecting cross-lingual interference patterns.}
  \label{tab:deepseek_en_bn_lang_stats}
\end{table*}

\begin{table*}[!ht]
  \centering
  \begin{adjustbox}{width=0.80\textwidth}
    \begin{tabular}{lccccc}
      \toprule
      \multicolumn{6}{c}{\textbf{Language presence in CoT solutions — \texttt{Qwen-3 8B} (Question: Bangla, CoT: Bangla)}} \\
      \midrule
      \multicolumn{3}{c}{\textbf{Aggregate language counts}} & \multicolumn{3}{c}{\textbf{Language combinations in CoT}} \\
      \cmidrule(lr){1-3} \cmidrule(lr){4-6}
      \textbf{Language} & \textbf{Count} & \textbf{Proportion (\%)} & \textbf{Combination} & \textbf{Count} & \textbf{Proportion (\%)} \\
      \midrule
      Bangla   & 1132 & 78.3 & Bangla + English           & 190 & 88.4 \\
      English  &  116 &  8.0 & Bangla + English + Greek   &   7 &  3.3 \\
      Chinese  &    0 &  0.0 & ---                        & --- & --- \\
      Mixed    &  197 & 13.6 & ---                        & --- & --- \\
      \bottomrule
    \end{tabular}
  \end{adjustbox}
  \caption{Overall language distribution in chain-of-thought (CoT) outputs generated by the \texttt{Qwen-3 8B} model when prompted with Bangla questions and instructed to reason in Bangla. The table reports the proportions of individual languages detected in CoT solutions alongside the observed mixed-language combinations. \texttt{Mixed} indicates responses incorporating tokens from multiple languages.}
  \label{tab:qwen3_bn_bn_lang_stats}
\end{table*}

\begin{table*}[!ht]
  \centering
  \begin{adjustbox}{width=0.80\textwidth}
    \begin{tabular}{lccccc}
      \toprule
      \multicolumn{6}{c}{\textbf{Language presence in CoT solutions — \texttt{Qwen-3 8B} (Question: English, CoT: Bangla)}} \\
      \midrule
      \multicolumn{3}{c}{\textbf{Aggregate language counts}} & \multicolumn{3}{c}{\textbf{Language combinations in CoT}} \\
      \cmidrule(lr){1-3} \cmidrule(lr){4-6}
      \textbf{Language} & \textbf{Count} & \textbf{Proportion (\%)} & \textbf{Combination} & \textbf{Count} & \textbf{Proportion (\%)} \\
      \midrule
      Bangla   & 1111 & 76.9 & Bangla + English            & 206 & 86.2 \\
      English  &  117 &  8.1 & Bangla + English + Greek    &  10 &  4.2 \\
      Chinese  &    0 &  0.0 & Bangla + Chinese + English  &   1 &  0.4 \\
      Mixed    &  217 & 15.0 & ---                         & --- & --- \\
      \bottomrule
    \end{tabular}
  \end{adjustbox}
  \caption{Aggregate language distribution in chain-of-thought (CoT) outputs for the \texttt{Qwen-3 8B} model when prompted with English questions and instructed to respond in Bangla. The table summarizes the overall presence and proportions of languages detected in CoT reasoning traces, alongside observed multilingual combinations. \texttt{Mixed} indicates CoTs containing tokens from multiple languages.}
  \label{tab:qwen_en_bn_lang_stats}
\end{table*}

\begin{table*}[!ht]
  \centering
  \begin{adjustbox}{width=0.80\textwidth}
    \begin{tabular}{lccccc}
      \toprule
      \multicolumn{6}{c}{\textbf{Language presence in CoT solutions — \texttt{Mathstral-7B} (Question: Bangla, CoT: Bangla)}} \\
      \midrule
      \multicolumn{3}{c}{\textbf{Aggregate language counts}} & \multicolumn{3}{c}{\textbf{Language combinations in CoT}} \\
      \cmidrule(lr){1-3} \cmidrule(lr){4-6}
      \textbf{Language} & \textbf{Count} & \textbf{Proportion (\%)} & \textbf{Combination} & \textbf{Count} & \textbf{Proportion (\%)} \\
      \midrule
      Bangla   & 610 & 42.1 & Bangla + English            & 812 & 99.4 \\
      English  &  20 &  1.4 & Bangla + English + Greek    &   3 &  0.4 \\
      Chinese  &   0 &  0.0 &                  &   --- &  --- \\
      Mixed    & 815 & 56.5 & ---                         & --- & --- \\
      \bottomrule
    \end{tabular}
  \end{adjustbox}
  \caption{Aggregate language distribution in chain-of-thought (CoT) outputs for the \texttt{Mathstral-7B} model when prompted with Bangla questions and instructed to respond in Bangla. The table reports overall counts and proportions of languages observed in CoT solutions, as well as the counts and percentages of mixed-language combinations. \texttt{Mixed} denotes CoTs containing tokens from more than one language.}
  \label{tab:mathstral_bn_bn_lang_stats}
\end{table*}

\begin{table*}[!ht]
  \centering
  \begin{adjustbox}{width=0.80\textwidth}
    \begin{tabular}{lccccc}
      \toprule
      \multicolumn{6}{c}{\textbf{Language presence in CoT solutions — \texttt{Mathstral-7B} (Question: English, CoT: Bangla)}} \\
      \midrule
      \multicolumn{3}{c}{\textbf{Aggregate language counts}} & \multicolumn{3}{c}{\textbf{Language combinations in CoT}} \\
      \cmidrule(lr){1-3} \cmidrule(lr){4-6}
      \textbf{Language} & \textbf{Count} & \textbf{Proportion (\%)} & \textbf{Combination} & \textbf{Count} & \textbf{Proportion (\%)} \\
      \midrule
      Bangla   & 598 & 41.4 & Bangla + English              & 819 & 99.5 \\
      English  &  24 &  1.7 & Bangla + English + Greek      &   2 &  0.2 \\
      Chinese  &   0 &  0.0 & Arabic + Bangla + English     &   1 &  0.1 \\
      Mixed    & 823 & 57.0 & Bangla + Cyrillic + English   &   1 &  0.1 \\
      \bottomrule
    \end{tabular}
  \end{adjustbox}
  \caption{Aggregate language distribution in chain-of-thought (CoT) outputs for the \texttt{Mathstral-7B} model when prompted with English questions and instructed to respond in Bangla. The table reports overall counts and proportions of languages observed in CoT solutions, as well as the counts and percentages of mixed-language combinations. \texttt{Mixed} denotes CoTs containing tokens from more than one language.}
  \label{tab:mathstral_en_bn_lang_stats}
\end{table*}



\end{document}